%% file: main.tex
\documentclass[11pt]{article}

\usepackage{microtype}
\usepackage{graphicx}
\usepackage{booktabs}
\usepackage{amsmath}
\usepackage{amsfonts}

\usepackage{amssymb}
\usepackage{bm}
\usepackage{enumitem}
\usepackage{xspace}
\usepackage{multirow}
\usepackage{float}
\usepackage{wrapfig}
\usepackage{ziplab-tech-report}
\usepackage{hyperref}

\newcommand{\reporttitle}{Any Time, Any Space: A World-State-Centric Framework for Agentic Video Generation}
\newcommand{\reportshorttitle}{Any Time, Any Space (ATAS)}
\newcommand{\reportauthors}{Weijie~Wang$^{1,*}$ \hspace{0.7em} Haomin~Zhang$^{1,*}$ \hspace{0.7em} Ye~Li$^{1,*}$ \hspace{0.7em} Akide~Liu$^{2}$ \hspace{0.7em} Zeyu~Zhang$^{3}$ \hspace{0.7em} Haoyu~Zhao$^{1}$ \hspace{0.7em} Yefei~He$^{1}$ \hspace{0.7em} Donny~Y.~Chen$^{2}$ \hspace{0.7em} Feng~Chen$^{4}$ \hspace{0.7em} Bohan~Zhuang$^{1,\dagger}$}
\newcommand{\reportaffiliations}{$^{1}$ Zhejiang University \quad $^{2}$ Monash University \quad $^{3}$ BAIR, UC Berkeley \quad $^{4}$ Adelaide University}
\newcommand{\reportdate}{October 13, 2026}

\newcommand{\reportcode}{\url{https://github.com/ziplab/ATAS}}
\newcommand{\reportproject}{\url{http://lhmd.top/atas}}
\newcommand{\reportemail}{\href{mailto:bohan.zhuang@zju.edu.cn}{\nolinkurl{bohan.zhuang@zju.edu.cn}}}

\newcommand{\reportkeywords}{Long Video Generation, Agentic Framework, Multi-shot Video Generation}
\newcommand{\reportfootnote}{Equal contribution. $^{\dagger}$ Corresponding author.}
\newcommand{\reportpdfauthors}{Weijie Wang, Haomin Zhang, Ye Li, Akide Liu, Zeyu Zhang, Haoyu Zhao, Yefei He, Donny Y. Chen, Feng Chen, Bohan Zhuang}
\newcommand{\reportlogoheight}{0.44cm}

\techreportlabel{ZIP Lab Technical Report}
\techreportshorttitle{\reportshorttitle}
\techreportsetlogoheight{\reportlogoheight}
\techreportdate{\reportdate}
\techreportconference{}
\techreportcode{\reportcode}
\techreportproject{\reportproject}
\techreportemail{\reportemail}
\techreportorcid{}
\techreportkeywords{\reportkeywords}
\techreportfootnote{\reportfootnote}
\techreportwordmark{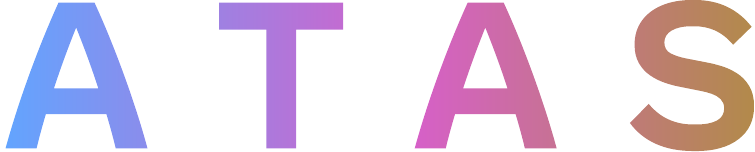}

\techreportlogos{%
  \raisebox{-0.5\height}{\includegraphics[viewport=0 0 176 176,clip,height=\reportlogoheight]{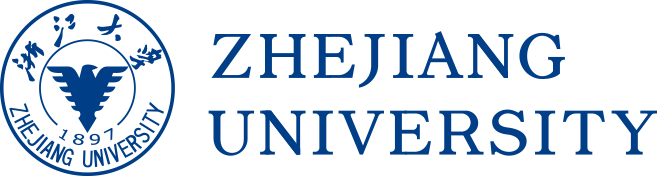}}\hspace{0.2em}%
  \raisebox{-0.5\height}{\includegraphics[viewport=208 20 657 145,clip,height=\reportlogoheight]{figures/logos/zju-logo-cropped.png}}\hspace{0.7em}
  \raisebox{-0.5\height}{\includegraphics[viewport=0 11 261 294,clip,height=\reportlogoheight]{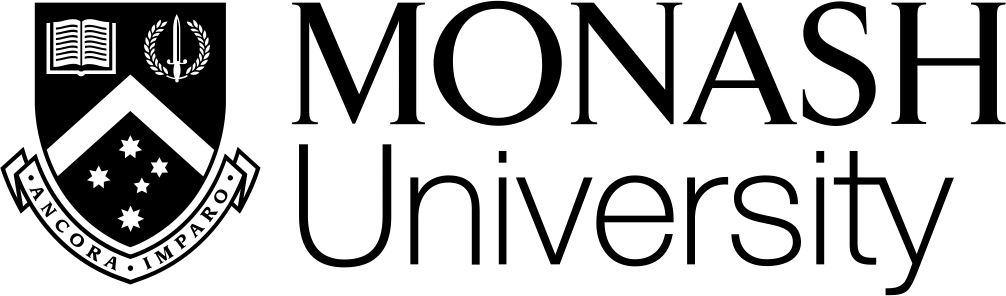}}\hspace{0.2em}%
  \raisebox{-0.5\height}{\includegraphics[viewport=291 0 1006 296,clip,height=\reportlogoheight]{figures/logos/monash-university-logo-cropped.png}}\hspace{0.7em}
  \raisebox{-0.5\height}{\includegraphics[viewport=0 0 260 260,clip,height=\reportlogoheight]{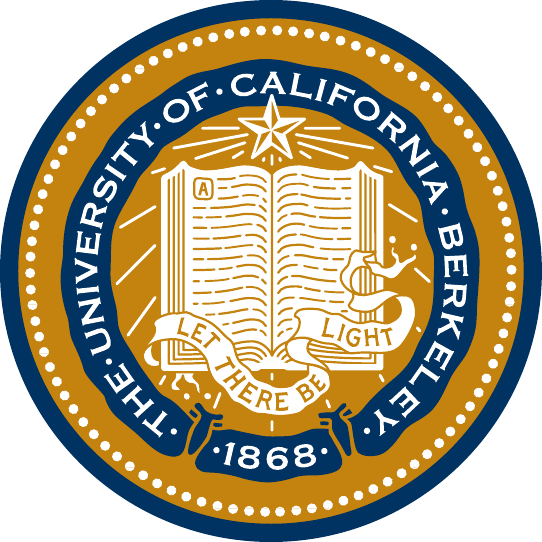}}\hspace{0.2em}%
  \raisebox{-0.5\height}{\includegraphics[viewport=313.44 24.21 997 234.59,clip,height=\reportlogoheight]{figures/logos/berkeley-horizontal.pdf}}\hspace{0.7em}
  \raisebox{-0.5\height}{\includegraphics[viewport=0 86.4 1290.24 817.92,clip,height=\reportlogoheight]{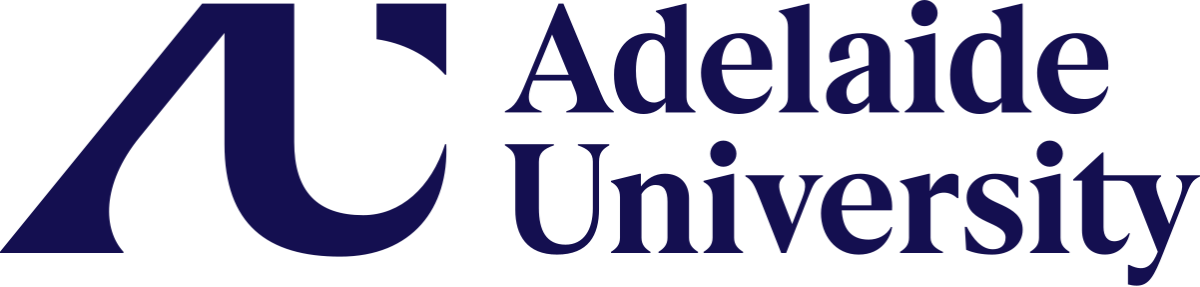}}\hspace{0.2em}%
  \raisebox{-0.5\height}{\includegraphics[viewport=1451.52 2.88 3456 826.56,clip,height=\reportlogoheight]{figures/logos/adelaide-logo.png}}\hspace{0.7em}
}

\hypersetup{
  colorlinks=true,
  linkcolor=ziplabblue,
  citecolor=ziplabblue,
  urlcolor=ziplabblue,
  pdftitle={\reporttitle},
  pdfauthor={\reportpdfauthors},
  pdfsubject={ZIP Lab technical report},
  pdfkeywords={\reportkeywords}
}

\newcommand{\method}{ATAS\xspace}

\newcommand{\sd}[1]{\,{\scriptsize$\pm$\,#1}}

\makeatletter
\newcommand{\atas@figuresfirst}[1]{%
  \begingroup
    \let\atas@figlist\@empty
    \let\atas@otherlist\@empty
    \def\@elt##1{%
      \@tempcnta\count##1\relax
      \divide\@tempcnta by 32\relax
      \ifnum\@tempcnta=1\relax
        \appto\atas@figlist{\@elt##1}%
      \else
        \appto\atas@otherlist{\@elt##1}%
      \fi}%
    #1%
    \let\@elt\relax
    \xdef#1{\atas@figlist\atas@otherlist}%
  \endgroup}
\pretocmd{\@cflt}{\atas@figuresfirst\@toplist}{}{\PackageError{ATAS}{Cannot order top floats}{}}
\pretocmd{\@vtryfc}{\atas@figuresfirst\@flsucceed}{}{\PackageError{ATAS}{Cannot order float pages}{}}

\makeatother
\newcommand{\boldstart}[1]{\par\addvspace{0.6\baselineskip}\noindent\textbf{#1}}
\begin{document}

\ziplabmaketitle{\reporttitle}{\reportauthors}{\reportaffiliations}{}
\ziplabendtitle{\input{sections/00_abstract}}

\begin{figure}[tp]
\centering
\includegraphics[width=\linewidth]{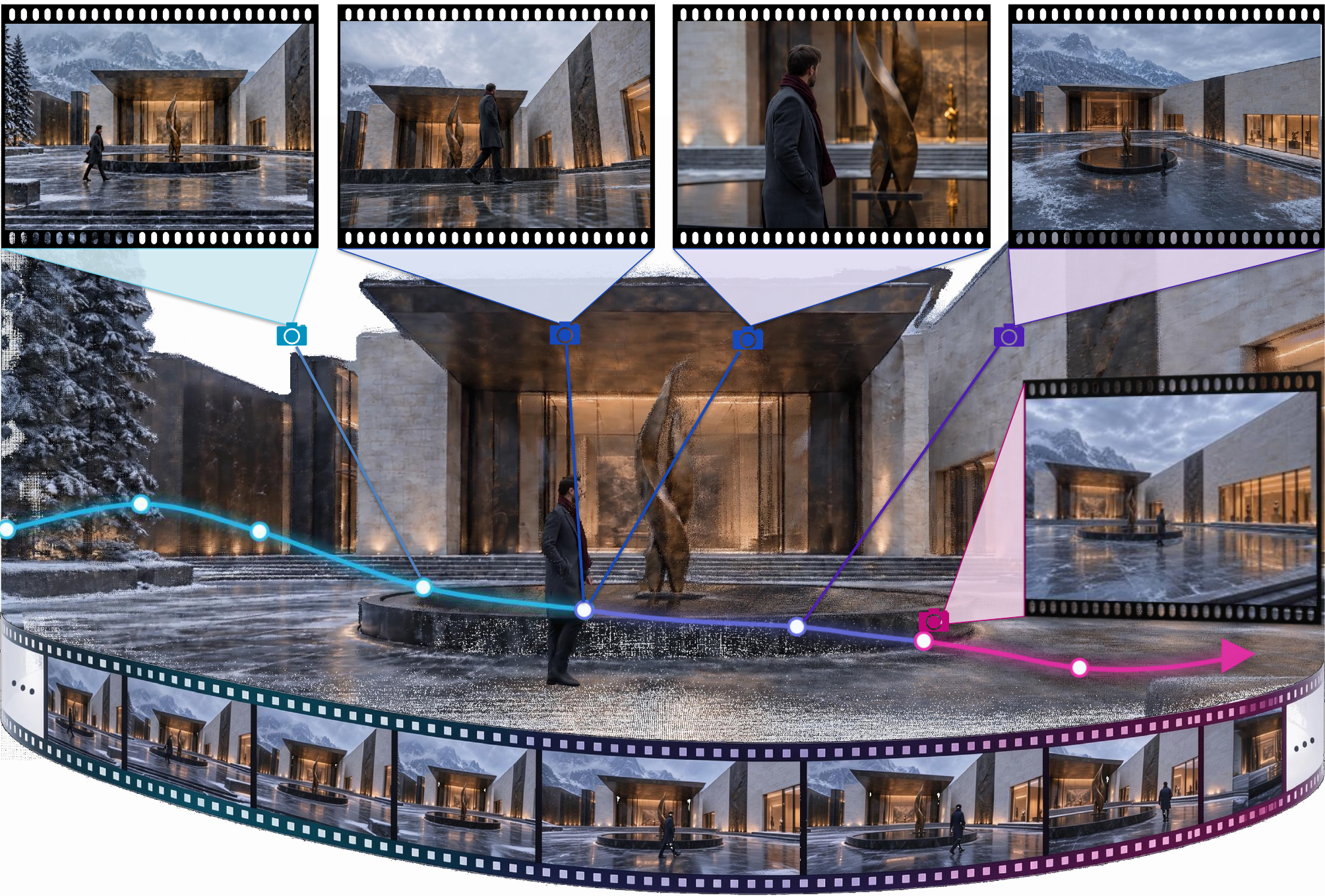}
\caption{\textbf{World-state-centric agentic video generation.} \method treats a multi-shot film as a sequence whose spatial and temporal information can be reused. Shared scene geometry anchors consistent return views, while time-indexed event evidence supports replay from a new camera. The generated shots form the visible film, and the resulting scene and event information guides subsequent requests.}
\label{fig:teaser}
\end{figure}

\input{sections/01_introduction}
\input{sections/02_related_works}
\input{sections/03_method}
\input{sections/04_experiments}
\input{sections/05_conclusion}
\input{sections/06_contribution_acknowledgements}

\FloatBarrier
\begingroup
\setlength{\bibsep}{0.5pt plus 0.5pt minus 0.5pt}\interlinepenalty=10000
\raggedright
\bibliographystyle{unsrtnat}
\bibliography{reference}
\endgroup

\clearpage
\appendix
\input{sections/appendix_contents}
\clearpage
\input{sections/06_appendix}

\end{document}

%% file: sections/00_abstract.tex
Long-form agentic video generation must turn a screenplay into shots that remain spatially and temporally consistent across nonconsecutive requests. Passing prompts or recent frames between shots does not provide a shared spatial frame for revisiting a scene or a time-indexed record for replaying an earlier event. We present \textbf{Any Time, Any Space (ATAS)}, which compiles a screenplay into shot requests with explicit location, camera, and event-time references. Before generation, it retrieves relevant evidence from a scene memory of registered static geometry and an event memory of dynamic observations indexed by scene, event, and time. After generation, ATAS reconstructs and registers the clip. It updates memory only when review and state checks accept the observation, while film delivery remains independent of memory admission. Across four video generators, ATAS improves all reported cross-shot consistency measures and achieves the highest aggregate benchmark score for each backbone. It also outperforms the compared agents and 3D-guided methods on target-image spatial revisit, leads three event-recapture criteria, and receives the highest human ratings on all six criteria.

%% file: sections/01_introduction.tex
\section{Introduction}
\label{sec:introduction}

Agentic video generation translates a screenplay into a sequence of video shots by coordinating shot planning, reference selection, and video synthesis. As a story unfolds across locations, times, and viewpoints, its coherence depends on preserving the spatial and temporal relationships established in earlier shots. For example, a story may depict an event in a room, shift to another location, and later return to the room or replay the earlier event from a different viewpoint. The return shot should preserve the established scene layout, while the replay should reproduce the same event, including its actions and temporal progression, under the new camera view. These requirements extend beyond the visual quality of individual clips~\citep{zhao2026worldolympiad}: they require scenes and events to remain consistently accessible across intervening shots, so that later generations remain grounded in a persistent underlying world, as illustrated in Fig.~\ref{fig:teaser}.

Existing agentic video systems typically preserve continuity through recent
frames, text summaries, or retrieved reference images. While effective for
locally consecutive shots, these cues neither organize past observations in a
shared spatial and temporal frame nor support non-local requests to revisit an
earlier scene or event. This motivates a persistent world state, but maintaining
such a state introduces two challenges. Past observations must remain
addressable in both space and time, and the state must evolve without blindly
absorbing generated content, whose visual, geometric, or temporal errors could
otherwise propagate to later shots. A practical system must therefore resolve
each request into a concrete shot specification, retrieve the relevant
spatial and temporal evidence, and incorporate the resulting observation into
the persistent world state only when it is sufficiently reliable.

To address these challenges, we introduce \textbf{Any Time, Any Space (ATAS)}, a world-state-centric framework for agentic video generation. ATAS maintains two complementary forms of persistent evidence: a scene state that preserves spatial structure in a shared coordinate frame, and an event state that retains dynamic observations with their temporal context. For each screenplay request, a world-state compiler grounds the requested shot in this world state, retrieves the relevant spatial and temporal evidence, and conditions a short-video generator on the resulting context.

Each generated clip is then treated as a candidate observation rather than trusted state. ATAS validates the reconstructed observation before incorporating it into the persistent world state, preventing unreliable generated evidence from propagating to subsequent shots. Importantly, whether a clip is suitable for the final video is separated from whether its observation is reliable enough to update the world state. This closed loop allows the world state to evolve across generation while remaining addressable for later spatial revisits and historical event recapture.


Our contributions are threefold:
\begin{itemize}[before=\interlinepenalty10000]
    \item We introduce ATAS, a world-state-centric framework that organizes screenplay-driven video generation around persistent scene and event memories, enabling later shots to query evidence from earlier observations in any space and any time.
    \item We develop complementary spatially registered scene and time-indexed event memories, together with a validation-controlled update mechanism that separates film acceptance from the admission of new evidence into persistent state.
    \item Experiments against five agent baselines across four generators show that ATAS improves every reported cross-shot consistency measure and attains the highest aggregate score in each backbone block. It also outperforms the evaluated agent and 3D-guided methods on all three spatial-revisit metrics, leads three of five event-recapture criteria, and receives the highest mean human rating on all six criteria.
\end{itemize}

%% file: sections/02_related_works.tex
\section{Related Work}

\boldstart{Video Generation.} Text-to-video and image-to-video models provide the renderer for long-video systems, improving appearance, motion, identity, and temporal coverage~\citep{ho2022videodiffusion,singer2022makeavideo,ho2022imagenvideo,hong2022cogvideo,bartal2024lumiere,kondratyuk2024videopoet,yang2024cogvideox,polyak2024moviegen,chen2024videocrafter2,videoldm2023}. Long-video methods extend temporal context through co-denoising, distributed generation, or frame-level direction~\citep{wang2023genlvideo,tan2024videoinfinity,hong2023direct2v}, while camera-, sparse-control-, and geometry-guided methods constrain view or motion~\citep{zhou2024storydiffusion,zhao2024moviedreamer,khachatryan2023text2videozero,xing2023dynamicrafter,guo2023sparsectrl,wu2025geometry,wang2026worldr1}. These models improve local rendering and short-term control, but do not specify how observations from multiple clips are registered, retained, or queried in a later shot.

\boldstart{Agentic Video Generation.} Agentic video systems add script parsing, shot planning, reference selection, and review to the generation loop~\citep{long2024videostudio,li2024animdirector,zheng2025videogen,wu2025movieagent,videodirectorgpt,dreamfactory,gong2023talecrafter,onestory2025,videomemory2026,han2025videobench}. Their inter-shot state is usually an editorial context of prompts, reference images, or generated histories. Such context can preserve local appearance without defining a shared coordinate frame, an event-time index, or a rule for admitting a generated observation into future state. Active view selection also supports spatial reasoning from 3D scene observations~\citep{zhao2026cov}. ATAS separates request compilation, memory retrieval, shot generation, and state update within the plan-act-observe loop of language agents~\citep{yao2023react}.

\boldstart{Explicit Memory for Video Generation.} Point-cloud and explicit-memory methods condition future generation on spatial evidence. Neural rendering and feed-forward reconstruction recover scene structure from sparse or multi-view observations~\citep{yu2021pixelnerfneuralradiancefields,chen2021mvsnerffastgeneralizableradiance,wang2025volsplat,shi2025depthrepresentations,wang2026trisplat,wang2026feedforward}, and 3D Gaussian Splatting provides an efficient explicit scene representation~\citep{kerbl20233d}. Multi-view compression improves reconstruction scalability~\citep{wang2025zpressor}. Spatial-memory methods accumulate or query scene geometry~\citep{zhao2025spatia,wu2025video,wang2026lsm,wang2025evoworld}. Dynamic view-synthesis methods study camera-controlled reshooting~\citep{yu2024viewcrafter,ren2025gen3c,lin2026vista4d}. World-generation methods explore geometry-consistent or navigable scene construction~\citep{dai2025fantasyworld,zhu2025worldsplat,song2025worldforge,yu2025wonderworld,cao2026hyworld}. Related 3D generation research studies compact shape representations and efficient decoding~\citep{lin2026ziptok3d,cui2026block3d}. These methods provide geometric memory, dynamic view synthesis, or world assets, but do not expose the same shot-local observation layer, event-time addressing, and state-update decision interface. ATAS separates static scene memory from event-local dynamic evidence and evaluates an observation before incorporating it into persistent state.

%% file: sections/03_method.tex
\section{Method}
\label{sec:method}

\method turns a high-level screenplay into short video shots while keeping the spatial and temporal evidence needed by later shots. The central object is a persistent world state, queried before each shot and updated by observations that pass validation.

\subsection{Terms and Workflow}
\label{sec:formulation}

The terminology is as follows. A \emph{source observation} $o^{\mathrm{src}}$ is the input frame or clip that starts a scene. Registering this observation establishes the scene coordinate frame. The observation and the coordinate frame are different objects. The \emph{scene memory} $\mathcal{S}_t$ stores the accepted static geometry in that frame, while the \emph{event memory} $\mathcal{E}_t$ stores accepted dynamic observations together with their scene, event, and time information. The \emph{world state} is simply the pair of memories at a particular step,
\begin{equation}
    \mathcal{W}_t=(\mathcal{S}_t,\mathcal{E}_t).
    \label{eq:state}
\end{equation}
Equation~\ref{eq:state} defines the persistent state used in the method. Scene and event refer to the contents of the two memories, while world state denotes their current pair after the first $t$ requests.

\begin{figure}[tp]
    \centering
    \includegraphics[width=0.98\textwidth]{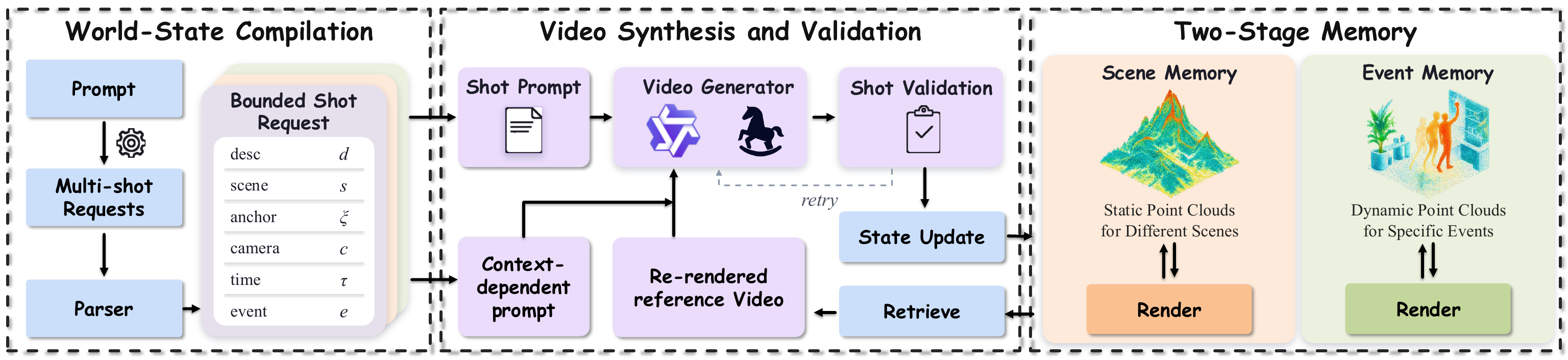}
    \caption{\textbf{Workflow of \method.} A high-level prompt is parsed into multi-shot requests and bounded shot requests. Each request contains a description, scene, spatial anchor, camera, time interval, and optional event. The shot prompt and the context-dependent prompt condition the video generator. Scene memory supplies static point-cloud references for the requested scene, while event memory supplies dynamic references for the requested event. Shot validation either requests another candidate or sends the accepted observation to the state update. The generated shot remains available as film independently of whether the observation is added to persistent state.}
    \label{fig:pipeline}
\end{figure}

The input is a high-level screenplay $u$, rather than a list of fully specified video prompts. At step $t$, the compiler produces a bounded shot request $q_t$, the memory query produces a conditioning context, and the generator produces a clip $V_t$. The clip is then converted into a candidate observation and either admitted to the next state or retained as film without a state update:
\begin{equation}
    (u,\mathcal{W}_t) \xrightarrow{\mathrm{Compile}} q_t
    \xrightarrow{\mathrm{Query}} C_t
    \xrightarrow{G} V_t
    \xrightarrow{\mathrm{Reconstruct/Register/Validate}}
    (V_t,\mathcal{W}_{t+1}).
    \label{eq:atas-transition}
\end{equation}
The next request uses the resulting state. Fig.~\ref{fig:pipeline} summarizes the workflow.

\subsection{World-state Compilation}
\label{sec:world_compilation}

Before generation, the compiler recursively splits the screenplay by scene, spatial and temporal relations, event references, and camera requirements until each request fits the short-video budget.

We write a renderable request as
\begin{equation}
    q_{t,k}=(d_{t,k},s_{t,k},\xi_{t,k},c_{t,k},\tau_{t,k},e_{t,k}),
    \label{eq:shot-request}
\end{equation}
where $d_{t,k}$ is the local shot description, $s_{t,k}$ identifies the scene, $\xi_{t,k}$ is the spatial anchor of the request in the scene coordinate frame, $c_{t,k}$ is the target camera, $\tau_{t,k}$ is the requested time interval, and $e_{t,k}$ is an optional event identifier.

For a screenplay segment, the compiler produces a request schedule
\begin{equation}
    \mathcal{Q}_t=\operatorname{Compile}(u,\mathcal{W}_t)
    =\{q_{t,1},\ldots,q_{t,K_t}\},
    \label{eq:compile}
\end{equation}
where $K_t$ is the number of requests currently emitted. Requests are processed in order, and a validated $\mathcal{W}_{t+1}$ conditions the next request. A rejected observation remains in the film but is not state evidence.

\subsection{Two-Stage Memory: Scene and Event Memory}
\label{sec:scene_event_memory}

\boldstart{Scene memory.} The registered source observation $o^{\mathrm{src}}$ establishes the scene coordinate frame and initializes $\mathcal{S}_0$. Later accepted observations are registered to that frame and merged as static point-cloud evidence with dynamic regions excluded. A request recovers a previously observed space by rendering these coordinates at its spatial anchor $\xi_t$ and requested camera:
\begin{equation}
    C_t^{\mathrm{scene}}=\operatorname{Render}(\mathcal{S}_t,c_t),
    \label{eq:scene_query}
\end{equation}
where $C_t^{\mathrm{scene}}$ is the re-rendered scene reference supplied to the generator. Reconstruction happens only after generation, when the clip is considered for a future state update~\citep{zhao2025spatia,wu2025video,wang2025evoworld,wu2025geometry}.

\boldstart{Event memory.} The event memory keeps dynamic evidence separate from static geometry. It is a set of records,
\begin{equation}
    \mathcal{E}_t=\{\epsilon_i\}_{i=1}^{N_t},
    \qquad
    \epsilon_i=(s_i,e_i,b_i,\tau_i,\mathcal{D}_i),
    \label{eq:event_record}
\end{equation}
where $s_i$ is the scene identifier, $e_i$ is the event identifier, $b_i$ identifies the source observation, $\tau_i$ is the source time interval, and $\mathcal{D}_i$ contains the registered frame-level RGB-D, mask, and dynamic point-cloud evidence. If a request refers to event $e_t$ during interval $\tau_t$, the event context is retrieved directly from $\mathcal{E}_t$:
\begin{equation}
    C_t^{\mathrm{event}}=
    \begin{cases}
    \operatorname{Retrieve}(\mathcal{E}_t,s_t,e_t,\tau_t),& e_t\neq\varnothing,\\
    \varnothing,&\text{otherwise.}
    \end{cases}
    \label{eq:event_query}
\end{equation}
The result is empty when no compatible record exists. Equation~\ref{eq:event_query} is the event branch of the query that produces $C_t^{\mathrm{event}}$. The memory retains frame-local dynamic evidence rather than an identity-resolved motion model~\citep{ren2025gen3c,lin2026vista4d}. Scene and event references are combined into the context-dependent prompt, while reconstruction remains a post-generation operation.

\subsection{Shot Generation from Queried State}
\label{sec:generation}

The local description in $q_t$ forms the shot prompt, while the scene and event references form the context-dependent prompt shown in Fig.~\ref{fig:pipeline}:
\begin{equation}
    C_t=(C_t^{\mathrm{scene}},C_t^{\mathrm{event}}),
    \qquad V_t=G(q_t,C_t),
    \label{eq:generation}
\end{equation}
where $G$ is the short-video generator. The shot prompt supplies local intent and its place and time, while $C_t$ supplies scene and, when needed, event references. This interface is kept fixed across comparisons~\citep{zhou2024storydiffusion,zhao2024moviedreamer,henschel2025streamingt2v}.

\subsection{Observation Registration and State Update}
\label{sec:observation_update}

The generated clip enters shot validation, which produces two separate decisions. Film review returns $f_t=1$ when the clip is accepted for delivery. If it is accepted, reconstruction and registration produce a candidate observation:
\begin{equation}
    \widetilde{\mathcal{O}}_t=\operatorname{Reconstruct}(V_t),
    \qquad
    \mathcal{O}_t=\operatorname{Register}(\widetilde{\mathcal{O}}_t,\mathcal{S}_t).
    \label{eq:observation}
\end{equation}
The registration stage returns $r_t=1$ when the reconstruction aligns to the scene coordinate frame. For the source observation, a valid registration establishes that frame. If film review fails, the system retries or rejects the candidate. If registration fails, the clip remains usable as film but cannot update state. The decisions come from the respective film-review and registration stages~\citep{umeyama1991least,wang2025vggt,wang2025continuous,lin2026vista4d}.

The state update check returns $a_t=1$ after the film and registration checks pass and the candidate has valid provenance, scene scope, and parent-state version. For a candidate that passes this check, the proposed next state is
\begin{equation}
    \widehat{\mathcal{W}}_{t+1}=\operatorname{Update}(\mathcal{W}_t,\mathcal{O}_t).
    \label{eq:proposed_update}
\end{equation}
The actual transition is
\begin{equation}
    \mathcal{W}_{t+1}=\begin{cases}
    \widehat{\mathcal{W}}_{t+1},& f_t=r_t=a_t=1,\\
    \mathcal{W}_t,&\text{otherwise.}
    \end{cases}
    \label{eq:state_update}
\end{equation}
Equation~\ref{eq:state_update} makes the decision rule explicit: $f_t$ controls film delivery, $r_t$ registration, and $a_t$ admission to persistent state. Film and state update are independent outputs, so $V_t$ remains in the film when $f_t=1$ even if $r_t=0$ or $a_t=0$. In that case, the clip is unavailable as evidence for later requests.

\begin{figure}[tp]
\centering
\includegraphics[width=\linewidth]{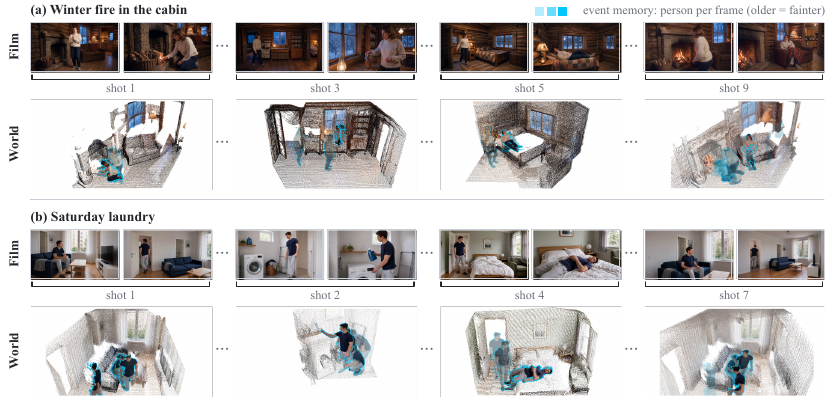}
\caption{\textbf{Qualitative world-state rollout.} Film frames and registered world views are shown for two story sequences. The examples illustrate how accepted observations extend the scene and event state across intervening shots.}
\label{fig:rollout}
\end{figure}

\subsection{Two Persistence Queries}
\label{sec:world_queries}

The same workflow supports two queries. For spatial revisit, the compiler emits the earlier scene identifier and the requested return anchor and camera. The system renders $\mathcal{S}_t$ and compares the generated return image with the target image. For historical event recapture, the request includes an earlier event identifier and time interval. The system retrieves the matching records from $\mathcal{E}_t$, combines them with the scene context, and generates the event from the requested camera. The first query tests recovery of static spatial evidence, and the second tests whether a time-local event remains addressable after intervening shots. The event protocol is source-grounded and does not assume a synchronized target video.

%% file: sections/04_experiments.tex
\begin{table}[tp]
\centering
\caption{\textbf{Agent-level quantitative evaluation on standard benchmarks.} Models are grouped by source type and rendering backbone. Methods within a block share the generator, prompt package, duration budget, and evaluation protocol. Bold and underlined entries mark the best and second-best values within each backbone block.}
\label{tab:main}
{
\setlength{\tabcolsep}{3.7pt}
\begin{tabular}{llccccc}
\toprule
\textbf{Model} & \textbf{Method} & \textbf{VBench} & \textbf{ST-Bench} & \shortstack{\textbf{ViStory-}\\\textbf{Self}} & \shortstack{\textbf{ViStory-}\\\textbf{Cross}} & \textbf{Avg.} \\
\midrule
\rowcolor{orange!20}\multicolumn{7}{c}{\textbf{Open-Source Models}} \\
\midrule
\multirow{6}{*}{\textsc{MiniMax-H3}} & I2V Extension & 0.828\sd{0.013} & 0.572\sd{0.020} & 0.601\sd{0.028} & 0.418\sd{0.027} & 0.605 \\
& VGoT & 0.834\sd{0.014} & 0.612\sd{0.017} & 0.638\sd{0.023} & 0.468\sd{0.030} & 0.638 \\
& Mora & 0.836\sd{0.012} & 0.598\sd{0.019} & 0.628\sd{0.026} & 0.452\sd{0.024} & 0.628 \\
& MovieAgent & \underline{0.852\sd{0.012}} & 0.608\sd{0.016} & 0.642\sd{0.025} & 0.476\sd{0.023} & 0.644 \\
& ReCA & 0.844\sd{0.011} & \underline{0.628\sd{0.018}} & \underline{0.658\sd{0.021}} & \underline{0.508\sd{0.022}} & \underline{0.659} \\
\rowcolor{green!12}\cellcolor{white}& \textbf{ATAS} & \textbf{0.860\sd{0.010}} & \textbf{0.691\sd{0.014}} & \textbf{0.710\sd{0.020}} & \textbf{0.547\sd{0.021}} & \textbf{0.702} \\
\midrule
\rowcolor{gray!15}\multicolumn{7}{c}{\textbf{Proprietary Models}} \\
\midrule
\multirow{6}{*}{\textsc{Wan~2.7}} & I2V Extension & 0.778\sd{0.014} & 0.486\sd{0.022} & 0.512\sd{0.029} & 0.348\sd{0.031} & 0.531 \\
& VGoT & 0.786\sd{0.011} & 0.536\sd{0.019} & 0.564\sd{0.026} & 0.412\sd{0.028} & 0.575 \\
& Mora & 0.791\sd{0.015} & 0.528\sd{0.017} & 0.558\sd{0.031} & 0.388\sd{0.033} & 0.566 \\
& MovieAgent & 0.808\sd{0.013} & 0.542\sd{0.021} & 0.572\sd{0.024} & 0.402\sd{0.026} & 0.581 \\
& ReCA & \textbf{0.854\sd{0.009}} & \underline{0.568\sd{0.018}} & \underline{0.598\sd{0.022}} & \underline{0.428\sd{0.029}} & \underline{0.612} \\
\rowcolor{green!12}\cellcolor{white}& \textbf{ATAS} & \underline{0.846\sd{0.012}} & \textbf{0.625\sd{0.016}} & \textbf{0.666\sd{0.021}} & \textbf{0.486\sd{0.032}} & \textbf{0.656} \\
\midrule
\multirow{6}{*}{\textsc{HappyHorse~1.0}} & I2V Extension & 0.836\sd{0.012} & 0.558\sd{0.023} & 0.592\sd{0.029} & 0.408\sd{0.031} & 0.599 \\
& VGoT & 0.822\sd{0.014} & 0.588\sd{0.018} & 0.628\sd{0.025} & 0.458\sd{0.027} & 0.624 \\
& Mora & 0.824\sd{0.011} & 0.574\sd{0.020} & 0.618\sd{0.027} & 0.442\sd{0.026} & 0.614 \\
& MovieAgent & \underline{0.856\sd{0.013}} & 0.598\sd{0.016} & 0.648\sd{0.021} & 0.472\sd{0.024} & 0.643 \\
& ReCA & 0.841\sd{0.010} & \underline{0.621\sd{0.019}} & \underline{0.668\sd{0.024}} & \underline{0.496\sd{0.025}} & \underline{0.656} \\
\rowcolor{green!12}\cellcolor{white}& \textbf{ATAS} & \textbf{0.862\sd{0.009}} & \textbf{0.668\sd{0.015}} & \textbf{0.706\sd{0.019}} & \textbf{0.540\sd{0.030}} & \textbf{0.694} \\
\midrule
\multirow{6}{*}{\textsc{Wan~3.0}} & I2V Extension & 0.836\sd{0.013} & 0.548\sd{0.021} & 0.582\sd{0.027} & 0.401\sd{0.030} & 0.592 \\
& VGoT & 0.834\sd{0.010} & 0.571\sd{0.019} & 0.622\sd{0.024} & 0.461\sd{0.026} & 0.622 \\
& Mora & 0.838\sd{0.012} & 0.562\sd{0.023} & 0.614\sd{0.022} & 0.438\sd{0.031} & 0.613 \\
& MovieAgent & \underline{0.858\sd{0.011}} & 0.578\sd{0.018} & 0.628\sd{0.025} & 0.452\sd{0.024} & 0.629 \\
& ReCA & 0.848\sd{0.009} & \underline{0.602\sd{0.017}} & \underline{0.661\sd{0.020}} & \underline{0.481\sd{0.027}} & \underline{0.648} \\
\rowcolor{green!12}\cellcolor{white}& \textbf{ATAS} & \textbf{0.877\sd{0.011}} & \textbf{0.719\sd{0.015}} & \textbf{0.756\sd{0.019}} & \textbf{0.584\sd{0.028}} & \textbf{0.734} \\
\bottomrule
\end{tabular}
}
\end{table}

\section{Experiments}
\label{sec:experiments}

\subsection{Experimental Setup}

\noindent\textbf{Standard benchmarks assess complementary aspects of long-form video generation.} We use the official inputs and evaluators for VBench~\citep{vbench_cvpr2024}, ST-Bench~\citep{zhang2025storymem}, and ViStory-Self and ViStory-Cross~\citep{vistorybench}. VBench measures aesthetics, subject and background consistency, temporal flicker, motion smoothness, and dynamic degree. ST-Bench evaluates aesthetics, global and single-shot prompt following, and cross-shot consistency with C-All and C-Top10. ViStory-Self measures identity, appearance, and temporal consistency within a story, while ViStory-Cross evaluates identity, appearance, and scene consistency across shots. Methods within each backbone block share the generator, prompt package, duration budget, and aggregation protocol. Per-metric scores appear in Appendix Tabs.~\ref{tab:app_vbench}--\ref{tab:app_vistory_cross}.

\noindent\textbf{Spatial revisit and event recapture use task-specific evidence.} Spatial revisit compares a generated return view with a target image under a common-visibility mask using PSNR$_{\mathrm C}$, SSIM$_{\mathrm C}$, and LPIPS$_{\mathrm C}$. The comparison includes multi-shot agent-based and 3D-guided view-synthesis methods. Event recapture uses a source-grounded VLM judge given sampled source-event frames, an event description and interval, the requested camera, and the generated clip. Its rubric scores event identity, action and state, temporal order, character and object consistency, and viewpoint realization. Binary decisions map to zero or one, ordinal ratings from one to five are normalized to zero through one, and scores are averaged across axes and then across event requests. No synchronized target video is used.

\noindent\textbf{Controlled analyses measure memory cost, component contributions, and human judgments.} Memory ablation fixes the ATAS agent and Wan~3.0 while varying stored evidence. Component ablation adds scene and event resolution, recursive compilation, observation validation, and state-update checks under matched requests. Efficiency measurements report per-story runtime, real-time factor, candidate-review outcomes, retries per shot, retry-adjusted delivery time, throughput, and memory overhead across six memory settings. Further analyses are in Appendix Tabs.~\ref{tab:app_memory_stats}--\ref{tab:app_update_check} and Figs.~\ref{fig:app_memory_sensitivity}--\ref{fig:app_update_check_tradeoff}. Human raters compare anonymized whole-video outputs from six methods on six criteria using a five-point scale.

\begin{figure}[tp]
\centering
\includegraphics[width=\textwidth]{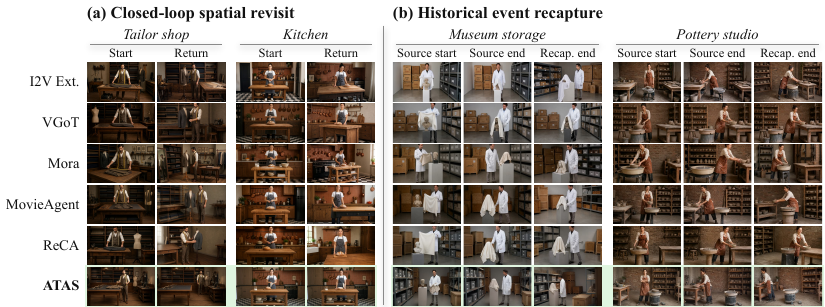}
\caption{\textbf{Spatial revisit and historical event recapture.} The top row uses a target image for spatial comparison. The bottom row uses source-grounded event evidence under a requested camera.}
\label{fig:capability_comparison}
\end{figure}

\begin{table}[tp]
\centering
\begin{minipage}[t]{0.44\linewidth}
\centering
\captionsetup{skip=4pt}
\captionof{table}{\textbf{Spatial revisit.} Agent-based and 3D-guided methods are evaluated against target images under a common-visibility mask, with ATAS shown separately.}
\label{tab:revisit}
{
\small
\setlength{\tabcolsep}{4pt}
\begin{tabular*}{\linewidth}{@{\extracolsep{\fill}}lccc}
\toprule
Method & PSNR$_{\mathrm C}$ & SSIM$_{\mathrm C}$ & LPIPS$_{\mathrm C}$ \\
\midrule
I2V Extension & 13.601 & 0.293 & 0.339 \\
VGoT & 14.444 & 0.313 & 0.300 \\
Mora & 13.796 & 0.318 & 0.288 \\
MovieAgent & 14.543 & 0.352 & 0.296 \\
ReCA & 13.870 & 0.268 & 0.306 \\
\midrule
FlexWorld & 14.740 & 0.343 & 0.315 \\
ViewCrafter & 15.255 & 0.375 & 0.293 \\
\midrule
\rowcolor{green!12}\textbf{ATAS} & \textbf{15.968} & \textbf{0.385} & \textbf{0.284} \\
\bottomrule
\end{tabular*}\par\vspace{0pt}
}
\end{minipage}\hfill
\begin{minipage}[t]{0.54\linewidth}
\centering
\captionsetup{skip=4pt}
\captionof{table}{\textbf{Source-grounded event recapture.} Five criteria assess event identity, action and state, temporal order, character and object consistency, and viewpoint realization against source-event evidence and the requested camera. No synchronized target video is used. Scores are averaged across criteria and event requests.}
\label{tab:recapture}
{
\small
\setlength{\tabcolsep}{1pt}
\renewcommand{\arraystretch}{1.05}
\begin{tabular*}{\linewidth}{@{\extracolsep{\fill}}lccccc}
\toprule
Method & \shortstack{Event\\id.} & \shortstack{Action\\state} & \shortstack{Temporal\\order} & \shortstack{Char.\\object} & \shortstack{View\\realization} \\
\midrule
I2V Extension & 0.450 & 0.433 & 0.433 & 0.817 & 0.200 \\
VGoT & 0.517 & 0.533 & 0.400 & 0.850 & 0.400 \\
Mora & 0.286 & 0.339 & 0.357 & 0.768 & 0.250 \\
MovieAgent & 0.500 & 0.450 & 0.500 & 0.783 & 0.350 \\
ReCA & 0.733 & \textbf{0.700} & 0.617 & \textbf{0.917} & 0.500 \\
\rowcolor{green!12}\textbf{ATAS} & \textbf{0.750} & 0.688 & \textbf{0.688} & \textbf{0.917} & \textbf{0.650} \\
\bottomrule
\end{tabular*}\par\vspace{0pt}%

}
\end{minipage}
\end{table}

\subsection{Main Results}

\noindent\textbf{Qualitative rollouts reveal continuity across nonconsecutive shots.} In both stories, the character returns to a previously visited place after intervening shots. The generated views retain evidence of the earlier scene layout and object state, while the reconstructed world views connect these returns to the spatial evidence accumulated during generation (Fig.~\ref{fig:rollout}). Further examples appear in Appendix Figs.~\ref{fig:app_qualitative_audit}--\ref{fig:app_story_grid} and Appendix~\ref{sec:app_gallery}.

\noindent\textbf{ATAS improves cross-shot consistency across generators while retaining competitive local quality.} In Tab.~\ref{tab:main}, it leads every reported aggregate on MiniMax-H3, Wan~3.0, and HappyHorse~1.0. With Wan~2.7 it remains ahead on ST-Bench and both ViStory consistency measures despite trailing on VBench. This pattern across open-source and proprietary backbones suggests that the gain comes from persistent multi-shot state rather than a single renderer. We compare with I2V Extension, VGoT~\citep{zheng2025videogen}, Mora~\citep{mora2024}, MovieAgent~\citep{wu2025movieagent}, and ReCA~\citep{liu2026reca} under matched backbone blocks.

\begin{figure}[tp]
\centering
\includegraphics[width=\linewidth]{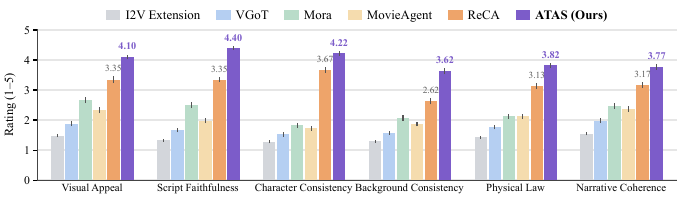}
\caption{\textbf{Human evaluation of agent methods.} Grouped bars show mean ratings and standard-error whiskers for six methods across six criteria.}
\label{fig:human_eval}
\end{figure}

\noindent\textbf{ATAS improves both target-view recovery and source-grounded event recapture.} The examples in Fig.~\ref{fig:capability_comparison} show return views and event replays from different cameras. ATAS reaches 15.968 PSNR$_{\mathrm C}$, 0.385 SSIM$_{\mathrm C}$, and 0.284 LPIPS$_{\mathrm C}$, outperforming the compared multi-shot and 3D-guided methods (Tab.~\ref{tab:revisit}). In Tab.~\ref{tab:recapture}, it leads event identity (0.750), temporal order (0.688), and viewpoint realization (0.650), ties ReCA on character and object consistency (0.917), and remains slightly lower on action-state preservation (0.688 versus 0.700). The remaining difficulty is exact action-state reproduction rather than retrieval of the requested occurrence or view.

\noindent\textbf{Human raters give ATAS the highest mean score on all six criteria.} Fig.~\ref{fig:human_eval} reports ratings for visual appeal, script faithfulness, character and background consistency, physical law, and narrative coherence under the anonymized protocol and criteria of~\citet{liu2026reca}. Ratings are highest for script faithfulness (4.40), character consistency (4.22), and visual appeal (4.10), while background consistency (3.62) and narrative coherence (3.77) remain lower. Participants therefore most clearly perceive faithful shot content and character carryover. Environment-wide consistency and whole-story coherence remain less resolved. Further details appear in Appendix~\ref{sec:app_human}.

\begin{figure}[tp]
\centering
\includegraphics[width=\linewidth]{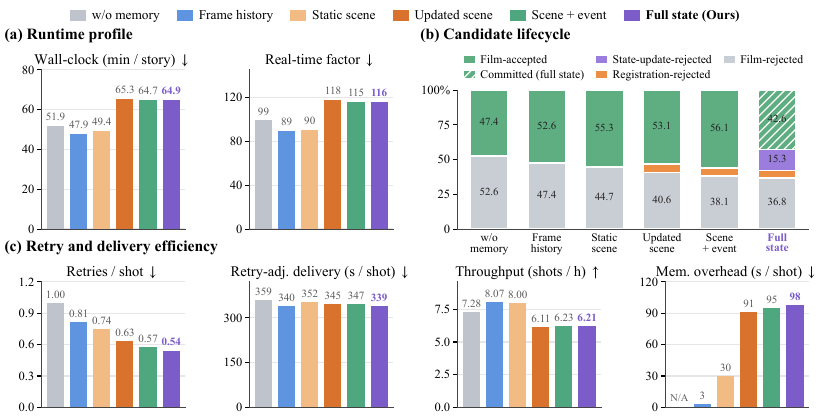}
\caption{\textbf{Efficiency comparison across memory settings.} Panels (a), (b), and (c) show runtime, candidate lifecycle, and retry and delivery measures across six memory settings.}
\label{fig:efficiency}
\end{figure}

\noindent\textbf{Persistent memory reduces retry demand at a measurable compute cost.} In Fig.~\ref{fig:efficiency}, retries fall from 1.00 to 0.54 per shot and retry-adjusted delivery time declines from 359 to 339 seconds per shot. Per-story runtime rises from 51.9 to 64.9 minutes, memory overhead reaches 98 seconds per shot, and throughput falls from 7.28 to 6.21 shots per hour. The benefit is fewer retries and faster retry-adjusted delivery, rather than lower total compute.

\noindent\textbf{Film-review rejection decreases as memory evidence accumulates.} The candidate-lifecycle panel shows a monotonic decline from 52.6\% with no memory to 36.8\% with full state, with each intermediate setting following the same direction. This trend indicates that the improvement is not confined to the final configuration. It also measures whether candidates can be delivered, complementing scores that evaluate the selected outputs.

\begin{table}[t]
\centering
\caption{\textbf{Agent-component ablation.} All rows use Wan~3.0 and the same requests, cameras, candidate budget, and retry policy. Components are added cumulatively from shot rendering through memory resolution, recursive compilation, observation validation, and state-update checking.}
\label{tab:ablation}
{
\begin{tabular}{lcccc}
\toprule
Configuration & \shortstack{ViStory\\average $\uparrow$} & C-Top10 $\uparrow$ & LPIPS$_{\mathrm C}$ $\downarrow$ & \shortstack{Event Recapture\\Score $\uparrow$} \\
\midrule
ATAS w/ shot rendering & 0.478 & 0.528 & 0.341 & 0.441 \\
ATAS w/ scene and event resolution & 0.612 & 0.568 & 0.334 & 0.508 \\
ATAS w/ recursive compilation & 0.638 & 0.684 & 0.288 & 0.711 \\
ATAS w/ observation validation & 0.651 & 0.702 & 0.286 & 0.726 \\
\rowcolor{green!12}\textbf{ATAS w/ state update check} & \textbf{0.670} & \textbf{0.719} & \textbf{0.284} & \textbf{0.739} \\
\bottomrule
\end{tabular}
}
\end{table}
\FloatBarrier
\clearpage
\FloatBarrier
\subsection{Ablation Study and Experimental Analysis}

\leavevmode
\begin{wraptable}{r}{0.54\textwidth}
\vspace{-\intextsep}
\centering
\caption{\textbf{Memory analysis.} ATAS with Wan~3.0 is fixed. Rows vary stored evidence and state checks.}
\label{tab:memory_quality}
{
\small
\setlength{\tabcolsep}{2.5pt}
\begin{tabular}{lccc}
\toprule
Config. & C-Top10$\uparrow$ & LPIPS$_{\mathrm C}\downarrow$ & Recap.$\uparrow$ \\
\midrule
ATAS w/o memory & 0.536 & 0.341 & 0.452 \\
ATAS w/ frame history & 0.598 & 0.319 & 0.548 \\
ATAS w/ static scene & 0.617 & 0.289 & 0.583 \\
ATAS w/ updated scene & 0.658 & 0.286 & 0.585 \\
ATAS w/ scene + event & 0.703 & 0.285 & 0.728 \\
\rowcolor{green!12}\textbf{ATAS w/ full state} & \textbf{0.719} & \textbf{0.284} & \textbf{0.739} \\
\bottomrule
\end{tabular}%

}
\end{wraptable}
\noindent\textbf{Scene and event memory improve different forms of continuity.} Frame history improves all three measures over no memory. Adding static scene evidence lowers LPIPS$_{\mathrm C}$ by 0.030, and updating that scene raises C-Top10 by 0.041 with little change in event score. Event memory gives the largest recapture gain with a smaller spatial change. Across the full ablation, C-Top10 rises from 0.536 to 0.719, event score from 0.452 to 0.739, and LPIPS$_{\mathrm C}$ falls from 0.341 to 0.284.

The modest final increment from scene-plus-event to full state suggests that validation preserves the main capabilities rather than adding another retrieval effect (Tab.~\ref{tab:memory_quality}).

\noindent\textbf{Scene and event memory establish gains that recursive compilation amplifies.} Adding memory raises ViStory~average from 0.478 to 0.612, C-Top10 from 0.528 to 0.568, and event score from 0.441 to 0.508. Recursive compilation then raises the event score to 0.711 and C-Top10 to 0.684 (Tab.~\ref{tab:ablation}).

\noindent\textbf{Film delivery and state admission.} Some clips are retained in the film even when registration or the state update is rejected. This lets the sequence continue without using an unaligned observation to condition later requests. The status breakdown separates these delivered clips from observations admitted to memory. Read with the runtime profile, it shows that persistent evidence reduces repeated review cycles while adding processing overhead, rather than making total generation faster.

\FloatBarrier

\setcounter{topnumber}{2}
\setcounter{totalnumber}{2}

%% file: sections/05_conclusion.tex
\section{Conclusion}

We presented ATAS, which compiles screenplays into shot requests grounded in explicit scene, camera, and event-time references, then updates persistent state only from validated observations. Across four video generators, ATAS improves cross-shot consistency, spatial revisit, event recapture, and human ratings. The ablations show complementary contributions from scene memory, event memory, and recursive compilation, while the efficiency analysis shows fewer retries at added compute cost. By keeping locations and prior events addressable throughout a story, ATAS provides a foundation for agentic video tools in interactive filmmaking, episodic content creation, and previsualization.

%% file: sections/06_contribution_acknowledgements.tex
\section*{Contribution and Acknowledgements}

\noindent\textbf{Author contribution.} Weijie Wang, Haomin Zhang, and Ye Li contributed equally to this work. Weijie Wang led the project and manuscript writing, contributed to code implementation, and conducted the initial experimental validation. Haomin Zhang implemented the code and contributed to experimental validation. Ye Li contributed to experimental validation and developed the demo and project page. Akide Liu contributed to the experiments and efficiency improvements. Zeyu Zhang, Haoyu Zhao, and Yefei He contributed to screenplay design and manuscript writing. Donny Y. Chen and Feng Chen contributed ideas to the project and guided the experimental design. Bohan Zhuang provided ongoing feedback throughout the project and participated in weekly project meetings.

\noindent\textbf{Acknowledgements.} We thank the participants in our human evaluation for their time and feedback. We also thank the researchers and developers who released the models, benchmarks, and open-source tools used in this work.

%% file: sections/appendix_contents.tex
\pdfbookmark[0]{Appendix Contents}{appendix.contents}
\section*{Appendix Contents}
\begingroup
\newcommand{\appcontentsentry}[3]{%
  \par\addvspace{\ifnum#1=0 0.8\baselineskip\else 0.2\baselineskip\fi}%
  \noindent\hspace*{\ifnum#1=0 0pt\else 1.5em\fi}%
  {\ifnum#1=0\bfseries\fi
  \hyperref[#2]{\ref*{#2}\quad #3}\nobreak\dotfill\hyperref[#2]{\pageref*{#2}}}\par}
\appcontentsentry{0}{sec:app_eval}{Evaluation Protocols and Benchmark Details}
\appcontentsentry{1}{sec:app_diagnostic}{Source-grounded Diagnostic Evaluator}
\appcontentsentry{1}{sec:app_vbench}{VBench Dimensions}
\appcontentsentry{1}{sec:app_stbench}{ST-Bench Metrics}
\appcontentsentry{1}{sec:app_vistory}{ViStory Metrics}
\appcontentsentry{0}{sec:app_compiler}{World-state compiler and bounded shot requests}
\appcontentsentry{0}{sec:app_state}{Memory Construction and Query}
\appcontentsentry{0}{sec:app_sensitivity}{Sensitivity and Mechanism Analysis}
\appcontentsentry{0}{sec:app_human}{Human Evaluation}
\appcontentsentry{0}{sec:app_qualitative}{Qualitative Results}
\appcontentsentry{0}{sec:app_gallery}{Long-Horizon Result Gallery}
\appcontentsentry{1}{sec:app_gallery_laundry}{Saturday Laundry in the Flat}
\appcontentsentry{1}{sec:app_gallery_fire}{Winter Fire in the Cabin}
\appcontentsentry{1}{sec:app_gallery_bath}{Running a Bath on a Winter Evening}
\endgroup

%% file: sections/06_appendix.tex
\setcounter{table}{0} \setcounter{figure}{0}
\renewcommand{\thetable}{\Alph{table}}
\renewcommand{\thefigure}{\Alph{figure}}
\renewcommand{\thesection}{\Alph{section}}
\renewcommand{\theHtable}{appendix.\Alph{table}}
\renewcommand{\theHfigure}{appendix.\Alph{figure}}
\renewcommand{\theHsection}{appendix.\Alph{section}}
\setlength{\abovecaptionskip}{5pt}
\setlength{\belowcaptionskip}{0pt}

This appendix provides implementation details and supporting analyses for reproducing and interpreting ATAS. It covers screenplay compilation, scene and event memory, benchmark evaluation, controlled analyses, human judgments, and extended story rollouts.

\section{Evaluation Protocols and Benchmark Details}
\label{sec:app_eval}

For each standard benchmark, the aggregate score is accompanied by metric-level details rather than story- or sequence-level listings. All benchmark-specific tables use the same backbone and method grouping as Tab.~\ref{tab:main}. The entries are interpreted with the evaluator definitions given in the corresponding subsection.

These breakdowns serve two purposes. First, they show which aspects of generation change when the rendering backbone changes, so that a single aggregate cannot be mistaken for a universal improvement. Second, they separate local rendering quality from the cross-shot behavior that the persistent state is intended to support. The dimensions are therefore read within a benchmark and within a backbone block. They are not combined into a new training objective or used to tune the reported agent.

\subsection{Source-grounded Diagnostic Evaluator}
\label{sec:app_diagnostic}

The source-grounded diagnostic evaluator uses one fixed VLM judge and one versioned rubric for every method. The evaluator receives sampled source-event frames, the event description and interval, the target-camera request, an optional target-camera memory render, and sampled frames from the generated clip. It does not receive a synchronized target video. The memory render is treated as a requested-view condition rather than ground truth.

For each axis, the judge returns either a binary decision or an ordinal rating from 1 to 5. We map a binary decision to 0 or 1 and map an ordinal rating $r$ to $(r-1)/4$. The normalized score for axis $a$ is the mean over the rubric items for that axis. The Event Recapture Score is the arithmetic mean of the five normalized axes in Tab.~\ref{tab:recapture}: event identity, action and state preservation, temporal phase and order, character and object consistency, and viewpoint realization. We first average within each event request and then average across requests so that longer clips do not receive more weight.

The same rubric is used for all methods and memory configurations. A request without source-event evidence is excluded before aggregation. A request without a generated clip receives zero on the applicable axes. Human judgments on matched clips are used as a separate agreement analysis rather than as hidden calibration data.

The five axes intentionally distinguish what is recovered from the source event. Event identity asks whether the requested occurrence is the one being reproduced. Action and state evaluate what the participants and objects are doing. Temporal phase and order evaluate where the recapture lies within the source interval. Character and object consistency check persistent entities, and viewpoint realization checks the requested camera relation. A high score on one axis therefore does not imply a high score on the others. A clip can show the correct characters while depicting the wrong phase of the event, or preserve the action while failing to realize the requested view. This decomposition is why the aggregate is reported together with the per-axis values in Tab.~\ref{tab:recapture}.

The evaluator is source-grounded rather than pixel-grounded. It can judge whether the generated clip agrees with the supplied event evidence and camera request, but it cannot establish pixel-level agreement with an unobserved target video. The resulting score is appropriate for historical event access and view-conditioned recapture. It should not be read as a substitute for synchronized reconstruction when such a target is available.

\subsection{VBench Dimensions}
\label{sec:app_vbench}

\begin{table}[tp]
\centering
\caption{\textbf{VBench dimension breakdown.} Each row is one reported VBench dimension, and each method/backbone column uses the official VBench evaluator and prompt suite. Higher is better unless the official dimension defines the opposite direction.}
\label{tab:app_vbench}
{
\begin{tabular}{lcccc}
\toprule
Dimension & \shortstack{Wan~2.7\\ATAS} & \shortstack{Wan~3.0\\ATAS} & \shortstack{MiniMax-H3\\ATAS} & \shortstack{HappyHorse\\ATAS} \\
\midrule
Aesthetic quality & 0.6387 & 0.6731 & 0.6945 & 0.6792 \\
Subject consistency & 0.9481 & 0.9643 & 0.9576 & 0.9418 \\
Background consistency & 0.9554 & 0.9717 & 0.9680 & 0.9522 \\
Temporal flickering & 0.9932 & 0.9863 & 0.9847 & 0.9884 \\
Motion smoothness & 0.9888 & 0.9766 & 0.9714 & 0.9795 \\
Dynamic degree & 0.5488 & 0.6882 & 0.5842 & 0.6314 \\
\bottomrule
\end{tabular}
}
\end{table}

Tab.~\ref{tab:app_vbench} shows that the rendering backbones trade different intrinsic properties. Wan~3.0 is strongest on subject and background consistency and dynamic degree, while Wan~2.7 is stronger on flicker and motion smoothness and MiniMax-H3 is strongest on aesthetic quality. The spread is largest for dynamic degree (0.5488--0.6882), whereas the temporal-flickering and motion-smoothness values remain high for every backbone. Thus the advantage in the standard multi-shot results cannot be explained by one uniformly better renderer: the backbones begin with different local strengths. This breakdown also gives a useful control for the later analyses. A change in cross-shot consistency should be interpreted alongside these local properties, rather than as evidence that ATAS improves every frame-level dimension equally. The table therefore separates backbone-dependent rendering behavior from the cross-shot gains measured by ST-Bench and ViStory.

\subsection{ST-Bench Metrics}
\label{sec:app_stbench}

\begin{table}[tp]
\centering
\caption{\textbf{ST-Bench metric breakdown.} The table reports official scores for aesthetic quality, global and single-shot prompt following, all-pair cross-shot consistency, and top-ten-pair cross-shot consistency.}
\label{tab:app_stbench}
{
\begin{tabular}{lccccc}
\toprule
Metric & \shortstack{Wan~2.7\\ATAS} & \shortstack{Wan~3.0\\ATAS} & \shortstack{MiniMax-H3\\ATAS} & \shortstack{HappyHorse\\ATAS} & \shortstack{ReCA\\reference} \\
\midrule
Aesthetic & 0.6387 & 0.6731 & 0.6945 & 0.6792 & 0.6517 \\
PF-G & 0.2174 & 0.2415 & 0.2351 & 0.2289 & 0.2216 \\
PF-S & 0.2154 & 0.2362 & 0.2308 & 0.2243 & 0.2192 \\
C-All & 0.5977 & 0.6844 & 0.6529 & 0.6615 & 0.5751 \\
C-Top10 & 0.6253 & 0.7187 & 0.6908 & 0.6684 & 0.6015 \\
\bottomrule
\end{tabular}
}
\end{table}

Tab.~\ref{tab:app_stbench} is consistent with the aggregate comparison: the strongest ATAS columns are the cross-shot consistency measures, especially C-Top10. Prompt-following changes are smaller and more uniform than the consistency changes: for example, Wan~3.0 reaches 0.2415 on PF-G and 0.2362 on PF-S, while its C-All and C-Top10 values are 0.6844 and 0.7187. The gap between C-Top10 and C-All is positive for every ATAS backbone (0.0069--0.0379), so the two consistency summaries provide complementary checks rather than duplicate columns. Wan~3.0 gives the strongest overall cross-shot behavior, with MiniMax-H3 and HappyHorse retaining substantial gains over the reference column. The pattern is therefore more specific than a general prompt-following gain: it is consistent with state being queried at shot boundaries and helping preserve information across the sequence.

\subsection{ViStory Metrics}
\label{sec:app_vistory}

\begin{table}[tp]
\centering
\caption{\textbf{ViStory-Self metric breakdown.} Rows report the supplied within-story consistency dimensions and aggregate.}
\label{tab:app_vistory_self}
{
\begin{tabular}{lccccc}
\toprule
Metric & \shortstack{Wan~2.7\\ATAS} & \shortstack{Wan~3.0\\ATAS} & \shortstack{MiniMax-H3\\ATAS} & \shortstack{HappyHorse\\ATAS} & \shortstack{ReCA\\reference} \\
\midrule
Identity consistency & 0.5494 & 0.6617 & 0.6421 & 0.5974 & 0.5608 \\
Appearance consistency & 0.6836 & 0.8139 & 0.7476 & 0.7772 & 0.7034 \\
Temporal self-consistency & 0.7648 & 0.7908 & 0.7392 & 0.7425 & 0.7186 \\
Aggregate & 0.6659 & 0.7555 & 0.7096 & 0.7057 & 0.6609 \\
\bottomrule
\end{tabular}
}
\end{table}

Tab.~\ref{tab:app_vistory_self} shows that the gains are not confined to one identity score. Wan~3.0 has the strongest appearance and temporal self-consistency and the highest aggregate, while the other ATAS backbones remain above the reference aggregate. Temporal self-consistency is the highest-scoring dimension for Wan~2.7, whereas appearance consistency is highest for the other three ATAS columns. Identity and temporal consistency move in the same direction, with Wan~3.0 reaching 0.6617, 0.8139, and 0.7908, respectively. A method that merely improves texture or color would not be expected to improve all three within-story dimensions together. The result indicates that the queried state supports both within-story identity and appearance continuity, while leaving room for future improvements in identity preservation.

\begin{table}[tp]
\centering
\caption{\textbf{ViStory-Cross metric breakdown.} Rows report the supplied cross-shot consistency dimensions and aggregate.}
\label{tab:app_vistory_cross}
{
\begin{tabular}{lccccc}
\toprule
Metric & \shortstack{Wan~2.7\\ATAS} & \shortstack{Wan~3.0\\ATAS} & \shortstack{MiniMax-H3\\ATAS} & \shortstack{HappyHorse\\ATAS} & \shortstack{ReCA\\reference} \\
\midrule
Cross-shot identity & 0.5073 & 0.5992 & 0.5736 & 0.5428 & 0.4934 \\
Cross-shot appearance & 0.4871 & 0.5806 & 0.5392 & 0.5618 & 0.5013 \\
Cross-shot scene consistency & 0.4621 & 0.5729 & 0.5285 & 0.5138 & 0.4484 \\
Aggregate & 0.4855 & 0.5842 & 0.5471 & 0.5395 & 0.4810 \\
\bottomrule
\end{tabular}
}
\end{table}

Tab.~\ref{tab:app_vistory_cross} follows the same ordering as the main cross-shot results. Wan~3.0 leads identity, appearance, and scene consistency, and every ATAS backbone exceeds the reference aggregate in this table. Scene consistency is the most demanding of the three columns for each ATAS backbone, which is expected because it depends on preserving the shared environment rather than only the subject. The relative ordering is stable across identity, appearance, and scene consistency. No backbone obtains its aggregate advantage from a single isolated submetric. These dimensions provide a direct check that the behavior persists across different shots rather than only within individual clips.

\section{World-state compiler and bounded shot requests}
\label{sec:app_compiler}

The world-state compiler receives a high-level screenplay request and the current scene and event evidence. It recursively resolves scene transitions, temporal relations, historical-event references, and camera requirements until each bounded shot request fits the short-video generation budget. Each request provides a local shot description, an explicit scene coordinate and target camera, and a requested time interval with optional event information. The compiler does not render a video while an internal request is being expanded. It emits a bounded shot request and waits for its validated observation before deciding which state a subsequent request may use.

\begin{figure}[tp]
\centering
\includegraphics[width=\linewidth]{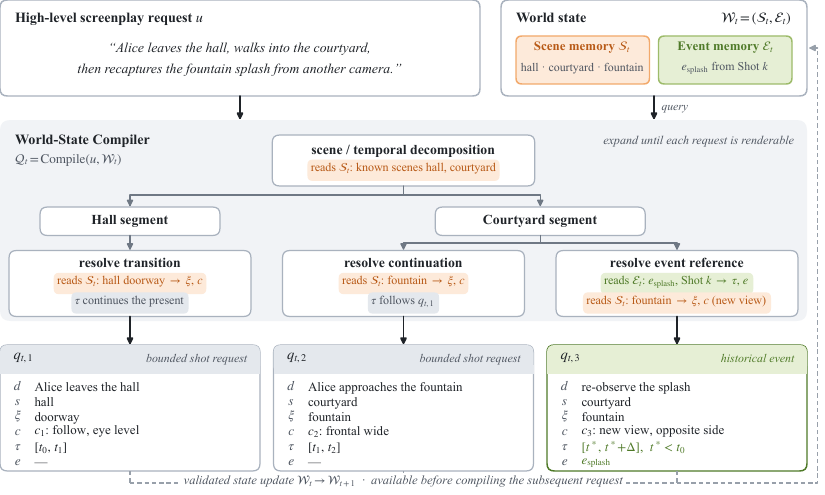}
\caption{\textbf{Recursive screenplay compilation.} The compiler turns a high-level request into bounded shot requests and uses validated state updates to condition subsequent requests.}
\label{fig:app_compiler}
\end{figure}

Fig.~\ref{fig:app_compiler} gives a concrete compiler trace from screenplay request to bounded shot requests and validated state updates.

The trace makes the unit of generation explicit. The screenplay is resolved into requests that carry a local description together with scene, camera, time, and optional event fields. These fields determine which evidence can be queried before generation and which evidence can be attached to the resulting observation afterward. A request that changes location can therefore establish a new scene relation, while a request that revisits an earlier location can reuse its scene coordinates. An event request additionally carries the time interval needed to select dynamic evidence. The figure checks that decomposition precedes rendering and that a later request consumes the validated state produced by an earlier one.

\section{Memory Construction and Query}
\label{sec:app_state}

Each generated clip is retained as a shot-local observation with its request, source provenance, timestamps, scene and event identifiers, target camera, reconstruction artifacts, and independent film, registration, and state-update statuses. The scene memory stores registered static evidence after dynamic regions are excluded. The event memory stores time-local dynamic evidence bound to a source shot, an event identifier, and a source interval. The two memories are queried separately and combined when a request refers to a historical event.

\begin{figure}[tp]
\centering
\includegraphics[width=\linewidth]{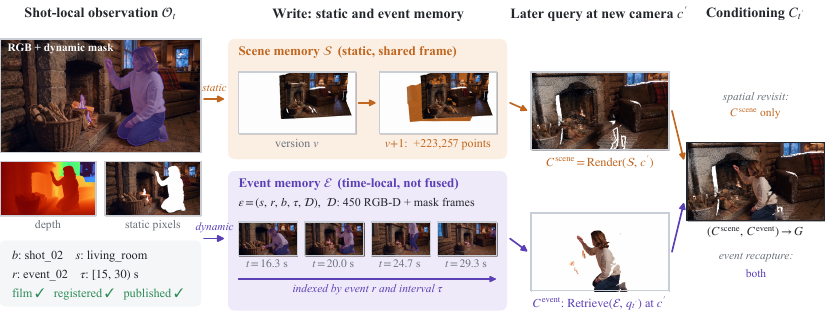}
\caption{\textbf{Scene and event memory.} The figure follows one accepted shot-local observation as static evidence is written to scene memory and dynamic evidence is indexed in event memory, then shows how both references condition a later camera request.}
\label{fig:app_memory_construction}
\end{figure}

Fig.~\ref{fig:app_memory_construction} connects each shot-local observation to the static and event query paths.

The separation is useful because the two evidence types have different persistence requirements. Static geometry is registered once to a scene coordinate frame and can be rendered at a new camera, whereas dynamic evidence must remain bound to an event and a time interval. Merging them into a single undifferentiated history would make a visually similar frame eligible even when it belongs to the wrong occurrence. The figure also clarifies the direction of information flow: reconstruction and registration create a candidate observation after generation, while rendering and event retrieval create conditioning references before the next generation. This distinction is central to interpreting the controlled memory rows in Tab.~\ref{tab:memory_quality}.

\begin{table}[tp]
\centering
\caption{\textbf{Memory statistics.} This analysis table reports memory growth and query cost at rollout checkpoints. It is not a sample manifest.}
\label{tab:app_memory_stats}
{
\begin{tabular}{lccccc}
\toprule
Checkpoint & \shortstack{Scene size\\($\times 10^{6}$)} & \shortstack{Event\\records} & \shortstack{Storage\\(GB)} & \shortstack{Query and\\update time (s)} & \shortstack{Processed\\shots} \\
\midrule
Initial state & 33.54 & 1.0 & 0.95 & 93.0 & 1.2 \\
After accepted update & 40.92 & 2.1 & 1.38 & 97.8 & 2.8 \\
End of rollout & 49.60 & 4.8 & 2.44 & 103.5 & 6.9 \\
\bottomrule
\end{tabular}
}
\end{table}

Tab.~\ref{tab:app_memory_stats} shows increasing scene and event coverage over the rollout while query and update time changes only modestly. From the initial state to the end of the rollout, scene size grows from 33.54 to 49.60 million units and event records from 1.0 to 4.8, while query and update time rises from 93.0 to 103.5. Storage increases from 0.95 to 2.44 in the reported units, so the state is not free. The cost grows more slowly than the amount of processed evidence in this short rollout. The processed-shot count also increases from 1.2 to 6.9, which makes the checkpoint comparison a measure of accumulation rather than a comparison of independent runs. These values should be read as an operating profile for the reported configuration, not as a general scaling law. They support treating the state as a growing evidence store whose cost is measured separately from the video generator.

\section{Sensitivity and Mechanism Analysis}
\label{sec:app_sensitivity}

The sensitivity analysis keeps the prompt package, camera requests, generator, candidate budget, and retry policy fixed while varying memory capacity, update frequency, memory freshness, and state update check strictness. The figures diagnose how state quality changes. They do not introduce a second benchmark.

\begin{figure}[tp]
\centering
\includegraphics[width=\linewidth]{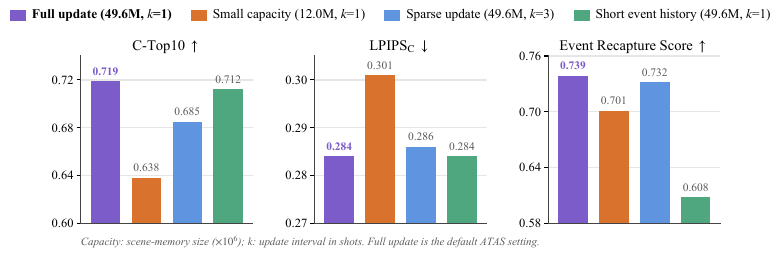}
\caption{\textbf{Memory capacity and update sensitivity.} The figure shows whether quality gains require additional state size or frequent updates.}
\label{fig:app_memory_sensitivity}
\end{figure}

\begin{table}[tp]
\centering
\caption{\textbf{Memory capacity and update sensitivity.} The table supplies the values used by Fig.~\ref{fig:app_memory_sensitivity}.}
\label{tab:app_memory_capacity}
{
\begin{tabular}{lccccc}
\toprule
Configuration & \shortstack{Capacity\\($\times 10^{6}$)} & \shortstack{Update\\interval} & C-Top10 & LPIPS$_{\mathrm C}$ & Event score \\
\midrule
Full update & 49.60 & 1 & 0.719 & 0.284 & 0.739 \\
Sparse update & 49.60 & 3 & 0.685 & 0.286 & 0.732 \\
Small capacity & 12.00 & 1 & 0.638 & 0.301 & 0.701 \\
Short event history & 49.60 & 1 & 0.712 & 0.284 & 0.608 \\
\bottomrule
\end{tabular}
}
\end{table}

Tab.~\ref{tab:app_memory_capacity} shows that sparse updates and reduced scene capacity lower both cross-shot consistency and event quality, while shortening the event history mainly affects the event score. Relative to full update, sparse update reduces C-Top10 from 0.719 to 0.685 and the event score from 0.739 to 0.732. Reducing capacity has a larger spatial effect (C-Top10 0.638 and LPIPS$_{\mathrm C}$ 0.301), while shortening event history reduces the event score to 0.608 with almost unchanged LPIPS$_{\mathrm C}$. Spatial quality therefore changes less than event quality when the event memory is shortened, which is consistent with the separation between static scene evidence and time-local event evidence. The comparison also suggests that update frequency and memory capacity are distinct controls: more frequent updates cannot fully compensate for a substantially smaller scene store, and a large scene store cannot replace missing event history.

\begin{figure}[tp]
\centering
\includegraphics[width=\linewidth]{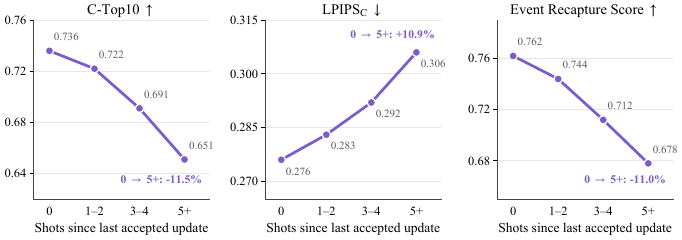}
\caption{\textbf{Memory freshness and staleness.} The figure tests whether quality degrades as future requests operate farther from the most recent accepted observation.}
\label{fig:app_staleness}
\end{figure}

\begin{table}[tp]
\centering
\caption{\textbf{Memory freshness data.} The table supplies the values used by Fig.~\ref{fig:app_staleness}.}
\label{tab:app_staleness}
\begin{tabular}{lccc}
\toprule
Shots since update & C-Top10 & LPIPS$_{\mathrm C}$ & \shortstack{Event Recapture\\Score} \\
\midrule
0 & 0.736 & 0.276 & 0.762 \\
1 to 2 & 0.722 & 0.283 & 0.744 \\
3 to 4 & 0.691 & 0.292 & 0.712 \\
5+ & 0.651 & 0.306 & 0.678 \\
\bottomrule
\end{tabular}
\end{table}

Tab.~\ref{tab:app_staleness} shows that quality declines as more shots are processed after the last accepted update. C-Top10 decreases from 0.736 at zero intervening shots to 0.651 after five or more, while LPIPS$_{\mathrm C}$ increases from 0.276 to 0.306. The event score follows the same direction, from 0.762 to 0.678. The gradual decline across all three measures is more informative than any single endpoint. It suggests that stale state gradually loses usefulness rather than at one fixed horizon. These bins do not identify whether the loss comes from scene changes, event density, or accumulated registration error. They motivate the separate state-update analysis below.

\begin{figure}[tp]
\centering
\includegraphics[width=\linewidth]{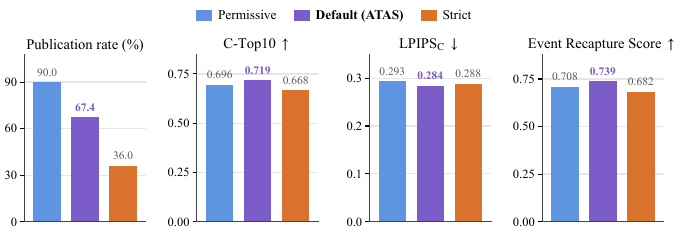}
\caption{\textbf{State update check trade-off.} The figure shows how stricter admission changes state coverage and downstream quality.}
\label{fig:app_update_check_tradeoff}
\end{figure}

Fig.~\ref{fig:app_update_check_tradeoff} visualizes the quality and state-coverage trade-off of the state update check.

\begin{table}[tp]
\centering
\caption{\textbf{State update check sensitivity.} The table supplies the values used by Fig.~\ref{fig:app_update_check_tradeoff}. It reports no independent benchmark.}
\label{tab:app_update_check}
\begin{tabular}{lcccc}
\toprule
\shortstack{State update\\setting} & \shortstack{State update\\rate} & C-Top10 & LPIPS$_{\mathrm C}$ & Event score \\
\midrule
Permissive & 90.0 & 0.696 & 0.293 & 0.708 \\
Default & 67.4 & 0.719 & 0.284 & 0.739 \\
Strict & 36.0 & 0.668 & 0.288 & 0.682 \\
\bottomrule
\end{tabular}
\end{table}

Tab.~\ref{tab:app_update_check} shows the quality-coverage trade-off of the state update check. A permissive setting admits 90.0\% of candidate updates but produces lower downstream quality (C-Top10 0.696 and event score 0.708), whereas a strict setting admits 36.0\% and also loses quality because fewer observations remain available (0.668 and 0.682). The default setting accepts 67.4\% and obtains the strongest downstream scores, 0.719 and 0.739, together with the lowest LPIPS$_{\mathrm C}$ of the three settings. This result does not imply that the default threshold is universally optimal. It shows that both indiscriminate admission and excessive rejection can reduce the evidence available to future requests. The check is therefore useful as a controlled admission mechanism, not as a replacement for improving reconstruction or retrieval.

\section{Human Evaluation}
\label{sec:app_human}
\setcounter{topnumber}{3}
\setcounter{totalnumber}{3}

The human study follows the grouped-bar comparison in Fig.~\ref{fig:human_eval}. Raters view anonymized clips in randomized order and score visual appeal, script faithfulness, character consistency, background consistency, physical law, and narrative coherence on a fixed five-point scale. The grouped bars report the six criteria, while this section reports score distributions and inter-rater agreement.

\begin{figure}[tp]
\centering
\includegraphics[width=\linewidth]{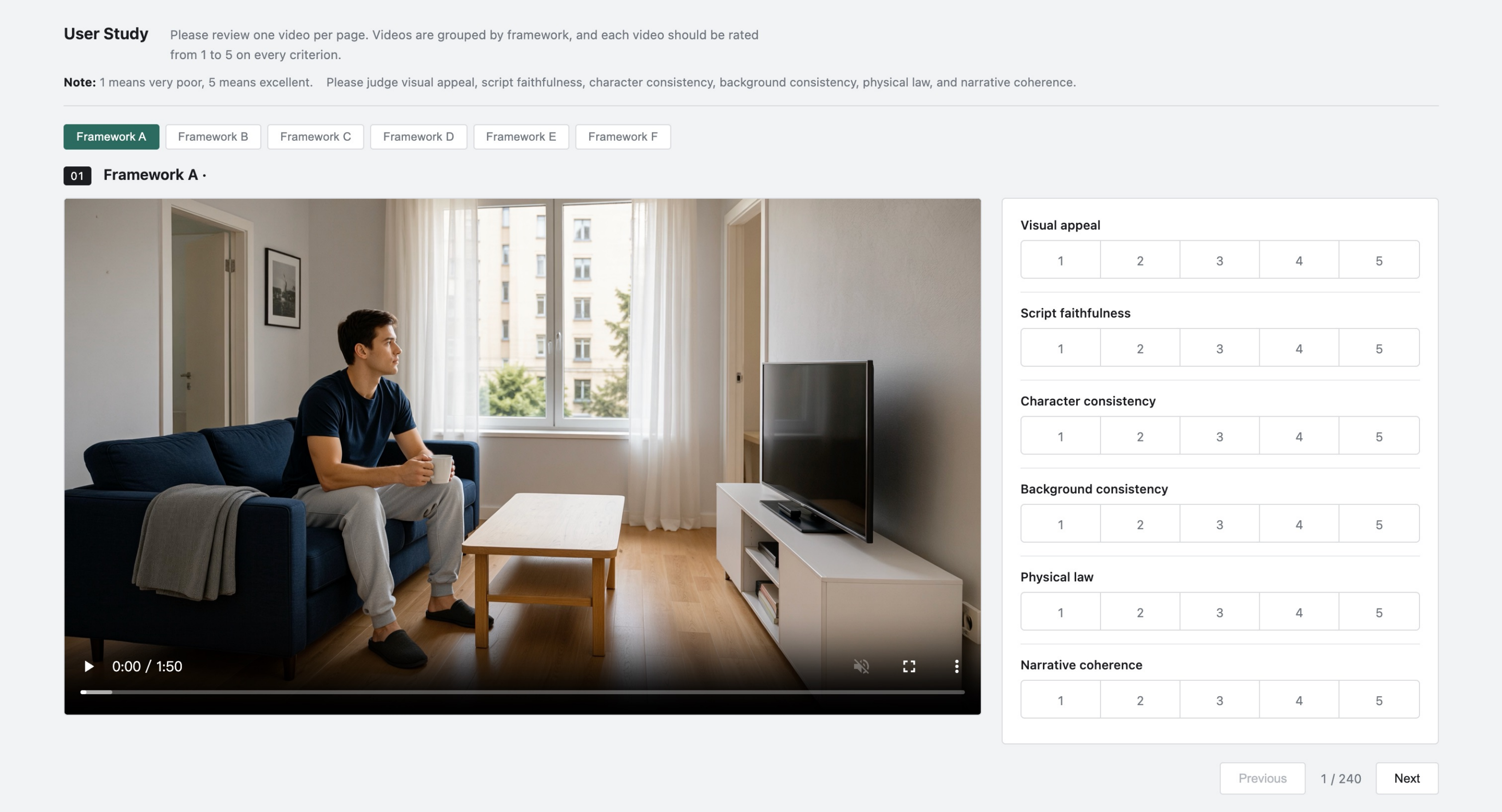}
\caption{\textbf{Human evaluation interface.} Each page shows one anonymized long video with its story title and synopsis. Raters watch the whole video and score it on the six criteria from 1 (very poor) to 5 (excellent). The method identity is hidden, and all six criteria must be rated before moving to the next video.}
\label{fig:app_user_study_interface}
\end{figure}

Fig.~\ref{fig:app_user_study_interface} shows the rating interface. Each video is rated as a whole rather than shot by shot, so the consistency criteria reflect the complete film.

\begin{figure}[tp]
\centering
\includegraphics[width=\linewidth]{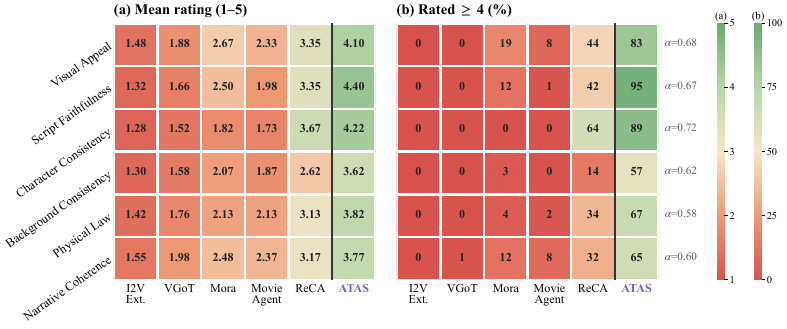}
\caption{\textbf{Distribution of human judgments.} The heatmaps report mean ratings and the fraction of ratings at least four for each criterion and method. The side panel summarizes inter-rater agreement.}
\label{fig:app_human_ratings}
\end{figure}

\begin{figure}[tp]
\centering
\includegraphics[width=\linewidth]{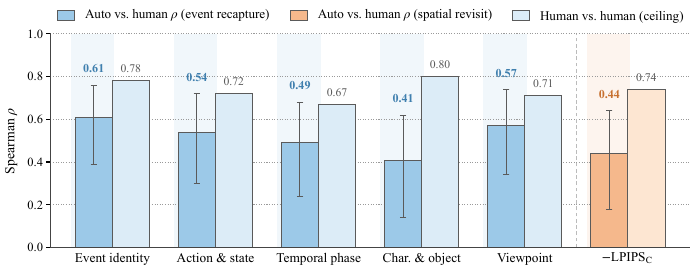}
\caption{\textbf{Automatic-to-human agreement.} Spearman correlations compare the event-recapture and spatial-revisit diagnostics with human ratings on matched clips, alongside the human--human agreement ceiling.}
\label{fig:app_human_agreement}
\end{figure}

Fig.~\ref{fig:app_human_ratings} shows score spread, and Fig.~\ref{fig:app_human_agreement} compares the automatic diagnostics with human judgments.

The human-rating heatmaps show a consistent method ordering across criteria. ATAS has the highest mean rating in all six rows, with means from 3.62 to 4.40. ReCA is the strongest baseline, with means from 2.62 to 3.67. The fraction of ratings at least four points makes the same distinction: ATAS ranges from 57\% to 95\%, whereas ReCA ranges from 14\% to 64\%. The agreement coefficients shown beside the heatmap range from 0.58 to 0.72, so the comparison is not based on a single criterion with unusually unstable ratings. These values support the qualitative conclusion that the benefit is visible to raters. The spread across criteria shows that narrative coherence and physical-law judgments remain harder than script faithfulness and character consistency.

The agreement plot provides a second check on the automatic diagnostics. Correlation with human ratings is 0.61, 0.54, 0.49, 0.41, and 0.57 for event identity, action and state, temporal phase, character and object, and viewpoint, respectively. The negated spatial distance, $-\mathrm{LPIPS}_{\mathrm C}$, correlates at 0.44; negation aligns its direction with higher human ratings. The corresponding human--human ceilings are higher (0.67--0.80), leaving room for evaluator noise and for aspects of perception that are not captured by the automatic scores. The result supports using the automatic metrics as structured diagnostics rather than treating them as complete substitutes for human judgment.

\section{Qualitative Results}
\label{sec:app_qualitative}

\begin{figure}[tp]
\centering
\includegraphics[width=\linewidth]{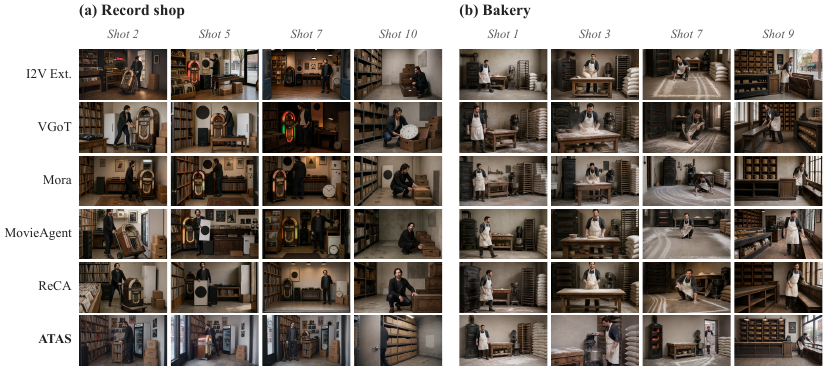}
\caption{\textbf{Qualitative baseline audit.} Matched frames from the record-shop and bakery stories compare methods across shot continuation, scene layout, object state, and character consistency.}
\label{fig:app_qualitative_audit}
\end{figure}

\begin{figure}[tp]
\centering
\includegraphics[width=\linewidth]{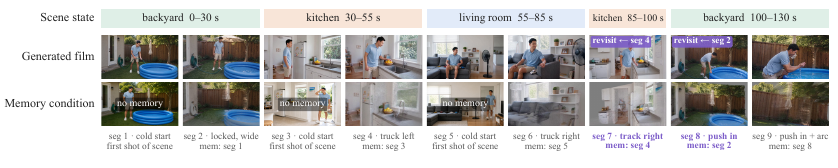}
\caption{\textbf{Representative rollout timeline.} The timeline follows the generated film and memory condition through scene changes, a spatial revisit, and an event recapture request.}
\label{fig:rollout_timeline}
\end{figure}

\begin{figure}[tp]
\centering
\includegraphics[width=\linewidth]{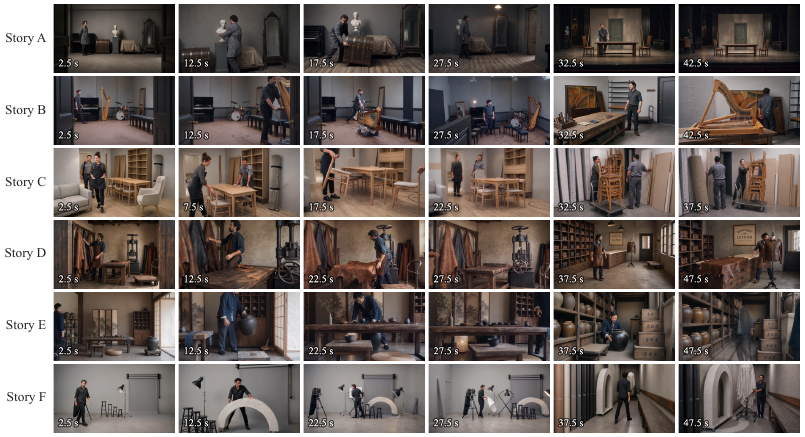}
\caption{\textbf{Representative multi-story results.} The grid shows timestamped frames from six stories, illustrating the range of scenes, actions, and camera requests used in the qualitative evaluation.}
\label{fig:app_story_grid}
\end{figure}

Figures~\ref{fig:app_qualitative_audit}--\ref{fig:app_story_grid} provide the qualitative audit, rollout timeline, and multi-story grid.

The baseline comparison highlights differences in shot-to-shot continuity. In the record-shop and bakery sequences, methods vary in whether room layout, object placement, and character state remain consistent after intervening shots. The ATAS row provides a reference for this behavior. A clip can remain plausible locally while drifting in layout or entity state, which is not captured by a single global similarity score. The rollout timeline shows when a scene state is created, when a later request queries it, and when a returned observation is written back. The six-story grid shows that these behaviors span different settings, people, objects, and camera patterns.

\setcounter{topnumber}{2}
\setcounter{totalnumber}{2}

\section{Long-Horizon Result Gallery}
\label{sec:app_gallery}
\newcounter{appfigsave}
\setcounter{appfigsave}{\value{figure}}
\setcounter{figure}{0}
\renewcommand{\thefigure}{\thesection.\arabic{figure}}
\renewcommand{\theHfigure}{gallery.\arabic{figure}}

The following three full-length \method films each visit three locations before returning to two of them. We examine whether the later views retain layouts, objects, and changes established before the intervening scenes. Each sequence includes an overview, five individual visits, and first-visit/return pairs. Frames come directly from the generated films, with timestamps in minutes and seconds. These examples show generated continuity rather than physical simulation of unobserved intervals.

\begin{figure}[tp]
\centering
\includegraphics[width=\linewidth]{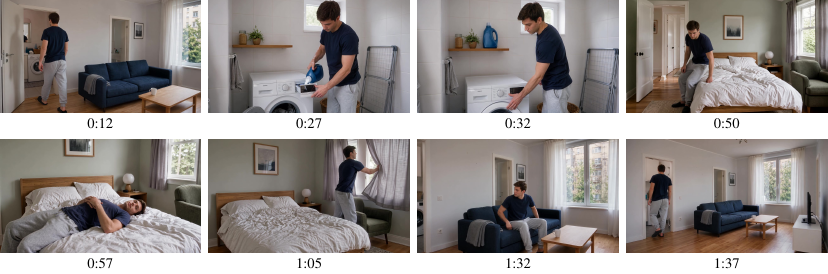}
\caption{\textbf{Saturday Laundry in the Flat: overview.} He leaves the living room for the laundry room, adds detergent and closes the washer, sits and then lies down on the bed, opens the curtains, and later gets up from the sofa to walk back to the laundry room.}
\label{fig:app_gallery_laundry}
\end{figure}

\subsection{Saturday Laundry in the Flat}
\label{sec:app_gallery_laundry}

A young man does his weekend laundry. The 1:50 film follows him from the living room to the laundry room and the bedroom, and then back to the living room and the laundry room. The story leaves two pieces of state behind: a mug set down on the coffee table, and a washer that is loaded and started before he leaves the laundry room.

\begin{figure}[tp]
\centering
\includegraphics[width=0.78\linewidth]{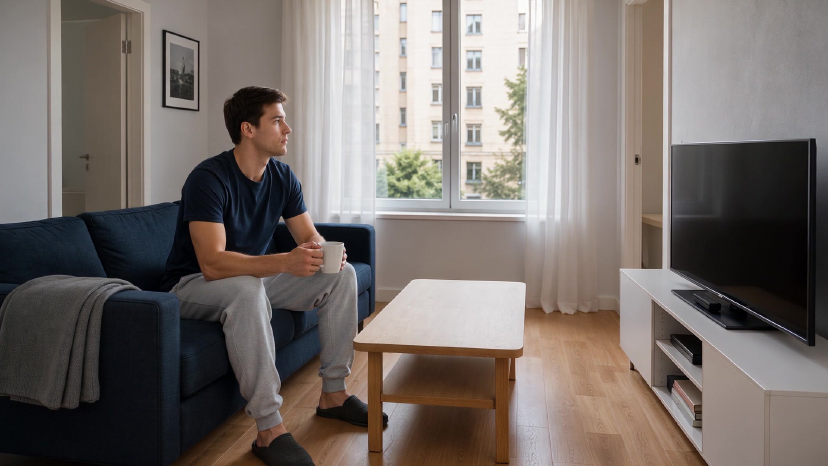}
\caption{\textbf{Saturday Laundry, visit 1: living room (0:02).} The first view of the living room, with the sofa, the gray throw, the coffee table, and the TV unit.}
\label{fig:app_gallery_laundry_v1}
\end{figure}

\begin{figure}[tp]
\centering
\includegraphics[width=0.78\linewidth]{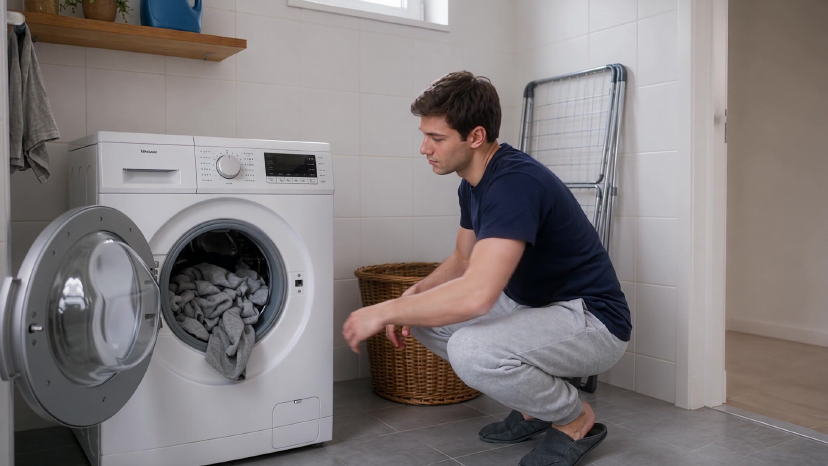}
\caption{\textbf{Saturday Laundry, visit 2: laundry room (0:22).} He loads the clothes from the basket into the washer.}
\label{fig:app_gallery_laundry_v2}
\end{figure}

\paragraph{Visit 1: living room (0:00--0:15).} The film opens on a quiet Saturday morning. He sits on the dark blue sofa with a mug of coffee, facing the TV, with a gray throw over the arm of the sofa and a low wooden coffee table in front of him. Before leaving, he sets the mug down on the table. This is the first observation of the living room, and the room is reconstructed from it together with the position of the mug.

\paragraph{Visit 2: laundry room (0:15--0:45).} He carries the basket into the small tiled laundry room, loads the clothes into the front-loading washer, adds detergent from the blue bottle on the shelf, and starts the machine. The shelf with the detergent, the plant, and the jar, the towel on the wall, and the folding drying rack by the door are established here. When he leaves, the washer is running and the basket is empty.

\begin{figure}[tp]
\centering
\includegraphics[width=0.78\linewidth]{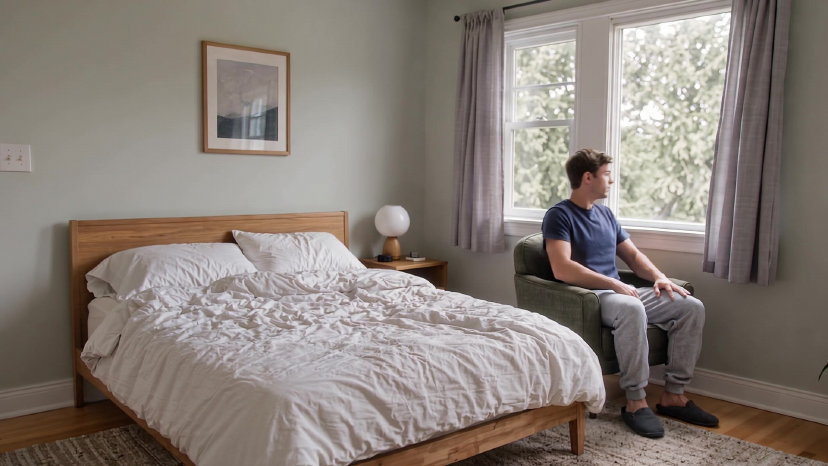}
\caption{\textbf{Saturday Laundry, visit 3: bedroom (1:07).} He sits in the armchair after opening the curtains.}
\label{fig:app_gallery_laundry_v3}
\end{figure}

\begin{figure}[tp]
\centering
\includegraphics[width=0.78\linewidth]{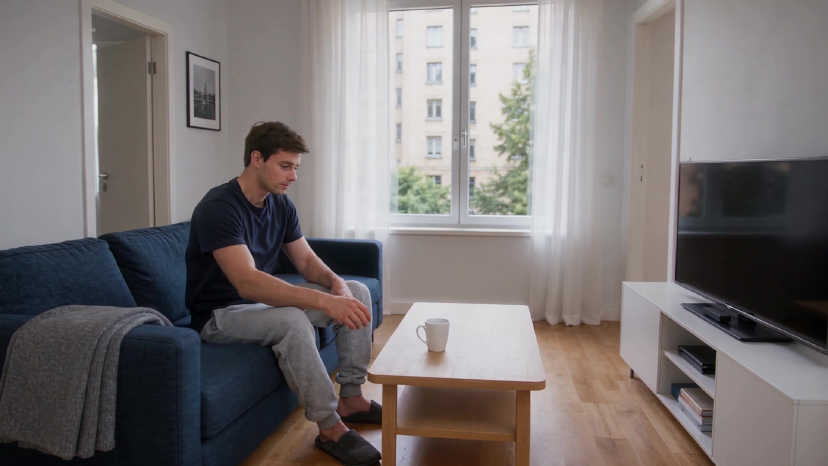}
\caption{\textbf{Saturday Laundry, visit 4: living room, return (1:25).} Back on the sofa, with the mug still on the coffee table where he set it down.}
\label{fig:app_gallery_laundry_v4}
\end{figure}

\paragraph{Visit 3: bedroom (0:45--1:15).} While the wash runs, he goes to the bedroom, lies down on the unmade bed, gets up to open the curtains, and sits in the green armchair by the window. The bedroom is seen only once in the film, so this visit only adds a new room to the world state. It also separates the two returns from the rooms in which the state was left.

\paragraph{Visit 4: living room, return (1:15--1:40).} He returns to the living room about a minute after leaving it. The stored living room conditions this visit, so the sofa, the throw, the coffee table, and the TV unit reappear in the same arrangement, and the mug he left behind is still standing on the table when he sits down again.

\begin{figure}[tp]
\centering
\includegraphics[width=0.78\linewidth]{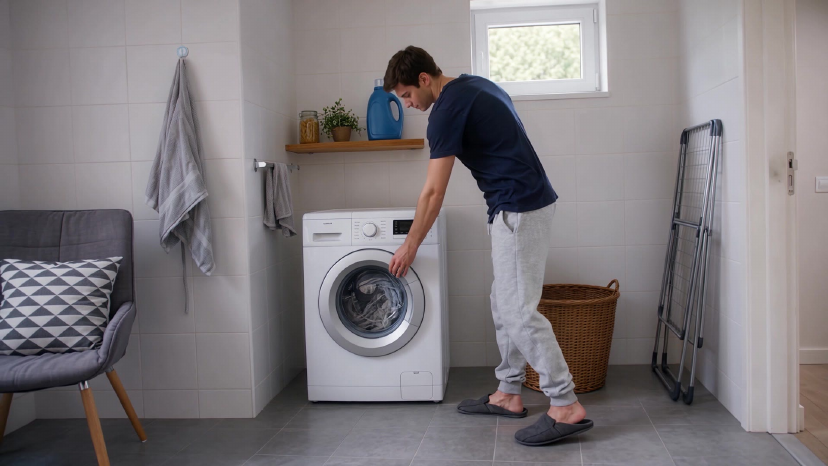}
\caption{\textbf{Saturday Laundry, visit 5: laundry room, return (1:45).} He opens the washer he started more than a minute earlier.}
\label{fig:app_gallery_laundry_v5}
\end{figure}

\begin{figure}[tp]
\centering
\includegraphics[width=0.8\linewidth]{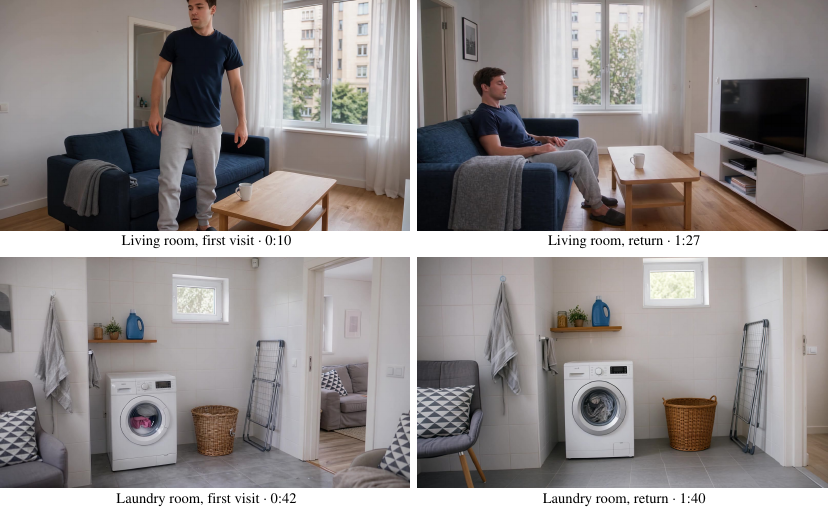}
\caption{\textbf{Saturday Laundry in the Flat: returns.} Top: the mug set down on the coffee table at the start is still there more than a minute later, next to the same sofa and throw. Bottom: the laundry room keeps its shelf, towel, basket, and drying rack, and the washer he started is still shut with the laundry inside.}
\label{fig:app_gallery_laundry_revisit}
\end{figure}

\paragraph{Visit 5: laundry room, return (1:40--1:50).} He goes back to collect the wash. The laundry room is rendered from its stored state after the wash was started: the shelf, the towel, the basket, and the drying rack are where they were, and the washer is still shut with the laundry inside. He opens the door of the finished machine.

\clearpage
\begin{figure}[tp]
\centering
\includegraphics[width=\linewidth]{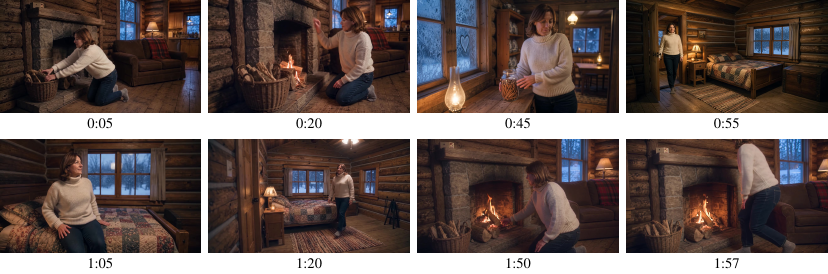}
\caption{\textbf{Winter Fire in the Cabin: overview.} She builds the fire from the log basket and watches it catch, stands by the frosted kitchen window with the heart she drew, rests in the bedroom, and returns to tend the fire, which has grown while she was away.}
\label{fig:app_gallery_fire}
\end{figure}

\begin{figure}[tp]
\centering
\includegraphics[width=0.78\linewidth]{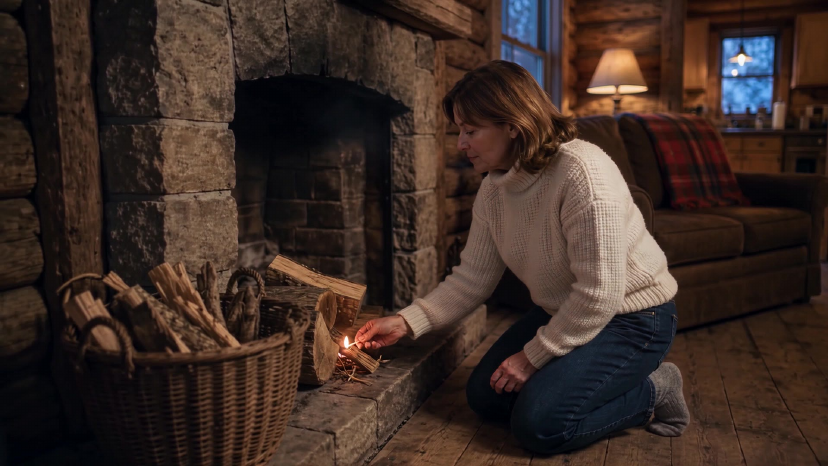}
\caption{\textbf{Winter Fire, visit 1: living room (0:12).} She lights the fire in the stone hearth.}
\label{fig:app_gallery_fire_v1}
\end{figure}

\subsection{Winter Fire in the Cabin}
\label{sec:app_gallery_fire}

A woman lights the fire in a cold log cabin. The 2:10 film moves from the living room to the kitchen and the bedroom, and then back through the kitchen to the living room. Both returns test state that she created herself: a heart drawn in the frost on the kitchen window, and a fire lit in the hearth before she leaves the living room.

\paragraph{Visit 1: living room (0:00--0:30).} She enters a cold living room with a stone fireplace, a log basket on the hearth, and a brown sofa with a red plaid throw. She kneels at the hearth, builds the fire from the logs in the basket, and lights it. When she leaves, a small fire is burning in a hearth that was cold at the start of the film.

\clearpage
\begin{figure}[tp]
\centering
\includegraphics[width=0.78\linewidth]{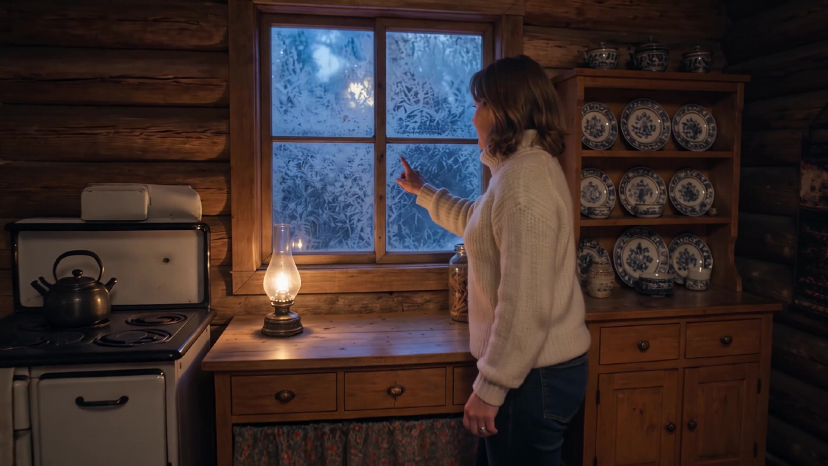}
\caption{\textbf{Winter Fire, visit 2: kitchen (0:35).} She draws in the frost on the kitchen window.}
\label{fig:app_gallery_fire_v2}
\end{figure}

\begin{figure}[tp]
\centering
\includegraphics[width=0.78\linewidth]{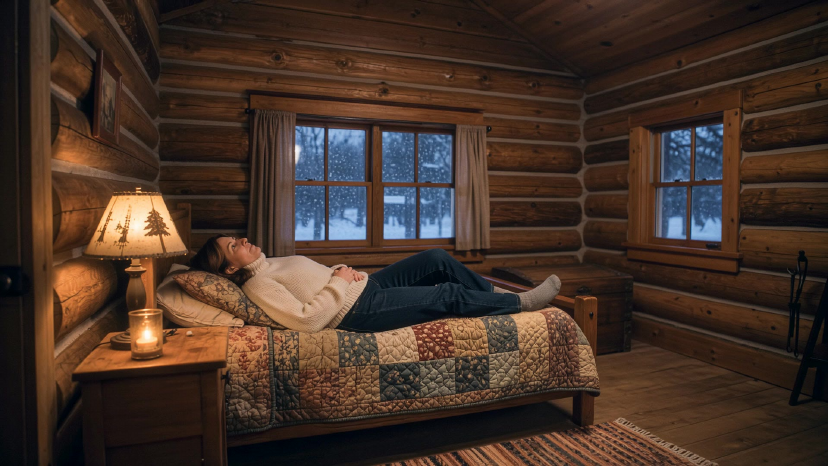}
\caption{\textbf{Winter Fire, visit 3: bedroom (1:10).} She lies on the bed under the lamp.}
\label{fig:app_gallery_fire_v3}
\end{figure}

\paragraph{Visit 2: kitchen (0:30--0:55).} In the kitchen, frost covers the window above the counter. She draws a heart in the frost with her finger, then turns to the jar and the oil lamp on the counter beside the dresser of blue-and-white plates. The heart is a small, specific mark that the kitchen must still carry when she comes back.

\paragraph{Visit 3: bedroom (0:55--1:25).} She rests in the bedroom, sitting on and then lying across the bed with the patchwork quilt, lit by the bedside lamp. The bedroom is seen only once. It adds a third room to the cabin and places half a minute of other content between the kitchen and its return.

\clearpage
\begin{figure}[tp]
\centering
\includegraphics[width=0.78\linewidth]{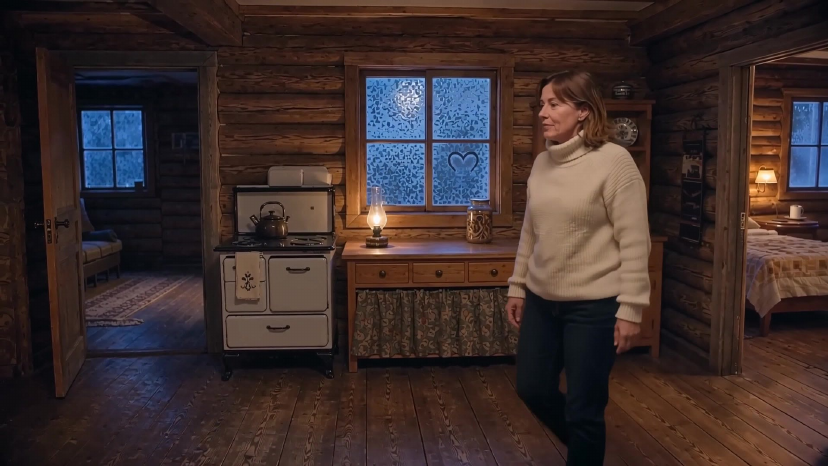}
\caption{\textbf{Winter Fire, visit 4: kitchen, return (1:30).} The kitchen as she left it, with the heart still in the frost on the window.}
\label{fig:app_gallery_fire_v4}
\end{figure}

\begin{figure}[tp]
\centering
\includegraphics[width=0.78\linewidth]{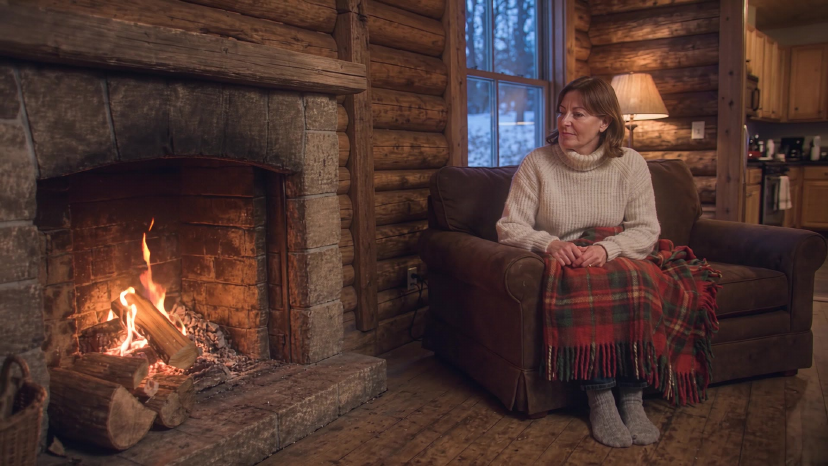}
\caption{\textbf{Winter Fire, visit 5: living room, return (2:05).} She sits by the fire she lit before leaving the room.}
\label{fig:app_gallery_fire_v5}
\end{figure}

\paragraph{Visit 4: kitchen, return (1:25--1:40).} On her way back she passes through the kitchen. The stored kitchen conditions this shot: the stove, the counter with the oil lamp, the dresser, and the doorways are in the same places, and the heart she drew is still visible in the lower right pane of the frosted window.

\paragraph{Visit 5: living room, return (1:40--2:10).} She returns to the living room about seventy seconds after lighting the fire. The fireplace, the log basket, the lamp, and the sofa with the red plaid throw match the first visit, and the fire she lit has kept burning and now fills the hearth. She settles into the armchair under a plaid blanket.

\clearpage
\begin{figure}[tp]
\centering
\includegraphics[width=0.8\linewidth]{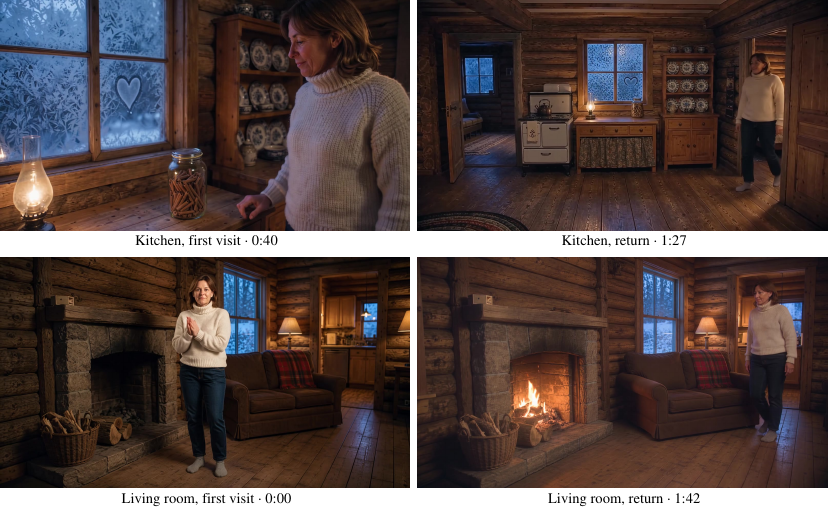}
\caption{\textbf{Winter Fire in the Cabin: returns.} Top: the heart she drew in the frost is still in the lower right pane of the same kitchen window. Bottom: the living room keeps its stone fireplace, log basket, lamp, and sofa with the red plaid throw, while the hearth that was cold at the start now holds the fire she lit.}
\label{fig:app_gallery_fire_revisit}
\end{figure}

\begin{figure}[tp]
\centering
\includegraphics[width=\linewidth]{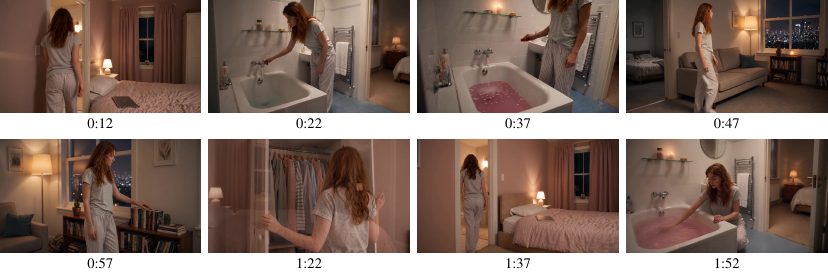}
\caption{\textbf{Running a Bath on a Winter Evening: overview.} She leaves the laptop on the bed, turns on the tap, lights a candle as the tub fills with pink water, browses the bookshelf in the living room, opens the wardrobe back in the bedroom, and tests the water in the filled tub.}
\label{fig:app_gallery_bath}
\end{figure}

\subsection{Running a Bath on a Winter Evening}
\label{sec:app_gallery_bath}

A young woman, tired from working on her laptop, runs herself a bath. The 2:10 film moves from the bedroom to the bathroom and the living room, and then back to the bedroom and the bathroom. The two returns test a static object left behind, the closed laptop on the bed, and a process that continues while she is away, the bath filling from the running tap.

\clearpage
\begin{figure}[tp]
\centering
\includegraphics[width=0.78\linewidth]{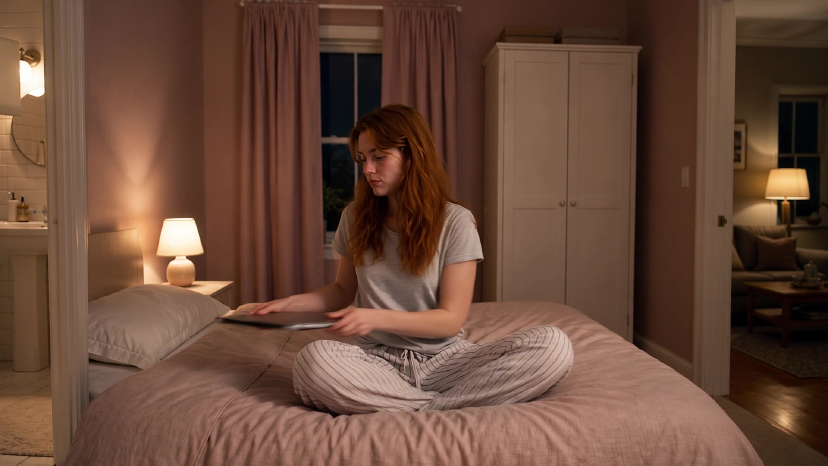}
\caption{\textbf{Running a Bath, visit 1: bedroom (0:07).} She closes the laptop on the bed.}
\label{fig:app_gallery_bath_v1}
\end{figure}

\begin{figure}[tp]
\centering
\includegraphics[width=0.78\linewidth]{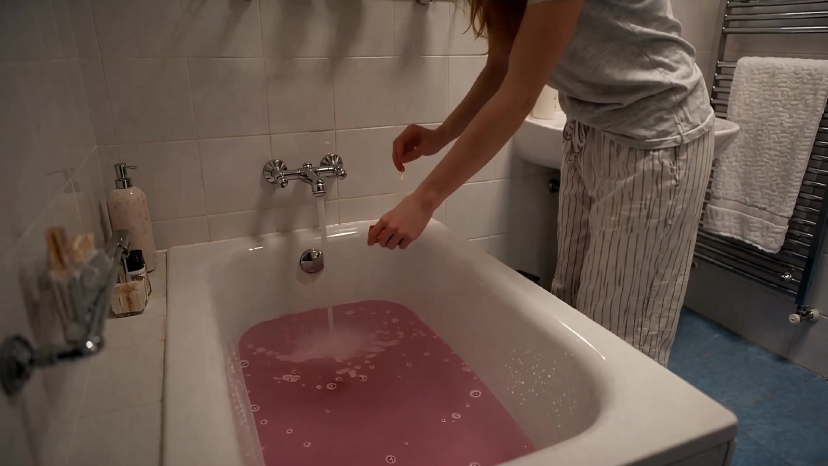}
\caption{\textbf{Running a Bath, visit 2: bathroom (0:30).} The tub begins to fill with pink water.}
\label{fig:app_gallery_bath_v2}
\end{figure}

\paragraph{Visit 1: bedroom (0:00--0:15).} She sits cross-legged on the bed in a warmly lit bedroom with pink curtains, a white wardrobe, and a lamp on the nightstand, working on her laptop. She closes the laptop, leaves it on the bed, and walks out through the door to the bathroom.

\paragraph{Visit 2: bathroom (0:15--0:45).} In the tiled bathroom she turns on the tap over the empty tub, pours in bath salts so that the water turns pink, and lights a candle on the glass shelf below the round mirror. She leaves with the tap running and the tub only partly filled.

\clearpage
\begin{figure}[tp]
\centering
\includegraphics[width=0.78\linewidth]{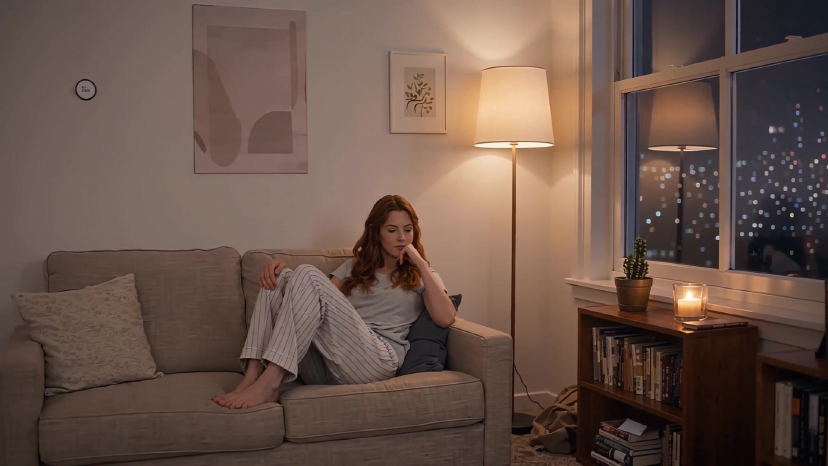}
\caption{\textbf{Running a Bath, visit 3: living room (1:07).} She rests on the sofa while the bath runs.}
\label{fig:app_gallery_bath_v3}
\end{figure}

\begin{figure}[tp]
\centering
\includegraphics[width=0.78\linewidth]{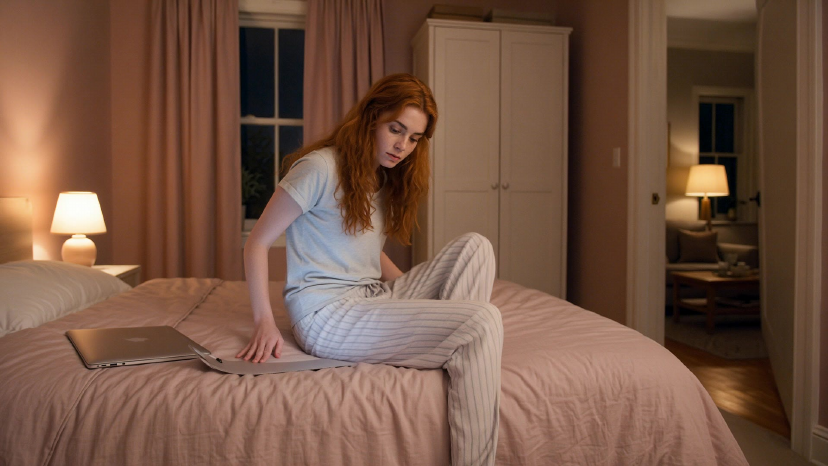}
\caption{\textbf{Running a Bath, visit 4: bedroom, return (1:30).} She sits next to the laptop she left on the bed.}
\label{fig:app_gallery_bath_v4}
\end{figure}

\paragraph{Visit 3: living room (0:45--1:15).} While the bath runs, she waits in the living room, looking out over the city lights, browsing the bookshelf, and resting on the sofa beside a small candle. The living room is seen only once, and it separates both returns from the rooms in which the state was left.

\paragraph{Visit 4: bedroom, return (1:15--1:40).} She comes back to the bedroom a minute after leaving it. From the same view as the opening shot, the bed, the nightstand lamp, the wardrobe, and the pink curtains match the first visit, and the laptop she closed is still lying on the bed. She sits beside it before heading to the bath.

\clearpage
\begin{figure}[tp]
\centering
\includegraphics[width=0.78\linewidth]{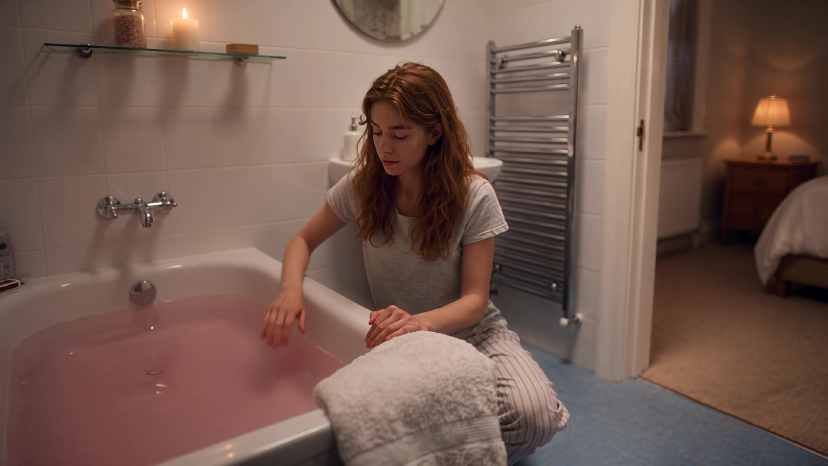}
\caption{\textbf{Running a Bath, visit 5: bathroom, return (2:05).} She kneels by the tub, which has filled while she was away.}
\label{fig:app_gallery_bath_v5}

\end{figure}

\begin{figure}[tp]
\centering
\includegraphics[width=0.8\linewidth]{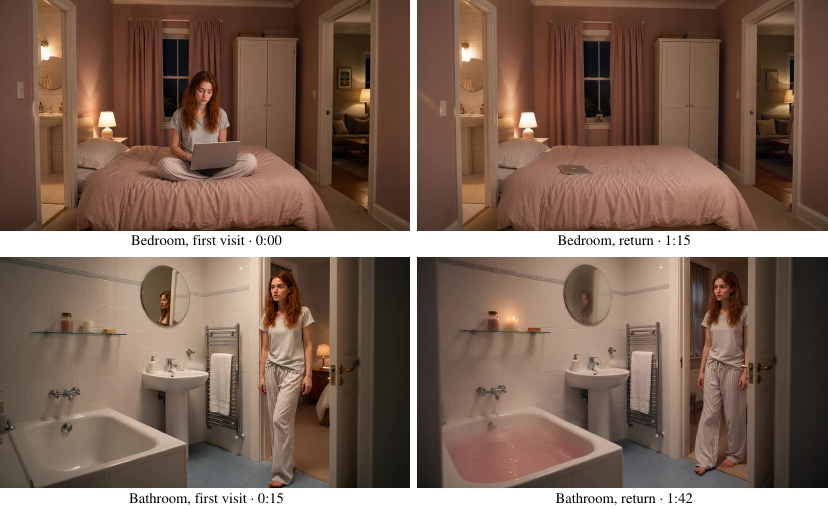}
\caption{\textbf{Running a Bath on a Winter Evening: returns.} Top: from the same view, the bedroom matches the first visit, with the bed, nightstand lamp, wardrobe, and pink curtains, and the laptop she closed is still lying on the bed. Bottom: the tiles, glass shelf, mirror, and towel rail match her first entrance, while the empty tub has filled with pink water and the candle she lit is still burning.}
\label{fig:app_gallery_bath_revisit}
\end{figure}

\paragraph{Visit 5: bathroom, return (1:40--2:10).} Back in the bathroom, the tiles, the glass shelf, the mirror, and the towel rail match her first entrance. The tub that was only partly filled when she left is now full of pink water, and the candle she lit is still burning. She kneels by the tub with a towel.